\documentclass{article}

\usepackage[final]{neurips_2023}

\usepackage[utf8]{inputenc} 
\usepackage[T1]{fontenc}    
\usepackage{hyperref}       
\usepackage{url}            
\usepackage{booktabs}       
\usepackage{amsfonts}       
\usepackage{nicefrac}       
\usepackage{xcolor}         

\usepackage{acronym}
\usepackage{multirow}
\usepackage{subcaption}
\usepackage{tocloft}

\usepackage{amsmath,amsfonts, amssymb}
\let\bm\boldsymbol  
\usepackage{upgreek}
\usepackage{mathtools}
\renewcommand{\vec}[1]{\bm{\MakeLowercase{#1}}}

\newcommand{\mat}[1]{\bm{\MakeUppercase{#1}}}

\newcommand{\reals}{\mathbb{R}}

\newcommand{\with}{\mathbin{;}}
\newcommand{\dx}[1][x]{\operatorname{d}\!#1}
\newcommand\der[3][]{\frac{\partial^{#1} {#2}}{\partial {#3}}}

\newcommand{\uniformdist}{\mathcal{U}}
\newcommand{\normaldist}{\mathcal{N}}

\DeclareMathOperator{\E}{\mathbb{E}}
\DeclareMathOperator{\Var}{Var}
\DeclareMathOperator{\Cov}{Cov}
\DeclareMathOperator{\Corr}{Corr}

\DeclareMathOperator{\kronecker}{\updelta}

\DeclareMathOperator{\softplus}{softplus}
\DeclareMathOperator{\relu}{ReLU}
\DeclareMathOperator{\leakyrelu}{LReLU}
\DeclareMathOperator{\elu}{ELU}

\DeclarePairedDelimiter\abs{\lvert}{\rvert}%
\DeclarePairedDelimiter\norm{\lVert}{\rVert}%

\newcommand{\stoptocwriting}{%
  \addtocontents{toc}{\protect\setcounter{tocdepth}{-5}}}
\newcommand{\resumetocwriting}{%
  \addtocontents{toc}{\protect\setcounter{tocdepth}{\arabic{tocdepth}}}}

\hypersetup{
   colorlinks=true,
   linkcolor=[RGB]{0, 132, 187},
   citecolor=[RGB]{0, 132, 187},
   urlcolor=[RGB]{0, 132, 187},
   pdfborder=0 0 0,
   pdftitle={},
   pdfsubject={}, 
   pdfkeywords={},
   pdfauthor={},%
   pdfstartview=FitH
}

\acrodef{iid}[i.i.d.]{identically and independently distributed}
\acrodef{ae}[AE]{Auto-Encoder}
\acrodef{icnn}[ICNN]{Input-Convex Neural Network}
\acrodef{cddd}[CDDD]{Continuous and Data-Driven Descriptor}

\title{Principled Weight Initialisation for \\ Input-Convex Neural Networks}

\stoptocwriting
\author{%
  Pieter-Jan Hoedt \& G\"{u}nter Klambauer \\
  LIT AI Lab \& ELLIS Unit Linz \\
  Institute for Machine Learning\\
  Johannes Kepler University, Linz, Austria \\
  \texttt{\{hoedt, klambauer\}@ml.jku.at}
}

\begin{document}

\maketitle

\begin{abstract}
    \acp{icnn} are networks that guarantee convexity in their input-output mapping. 
    These networks have been successfully applied for energy-based modelling, optimal transport problems and learning invariances.
    The convexity of \acp{icnn} is achieved by using non-decreasing convex activation functions and non-negative weights.
    Because of these peculiarities, previous initialisation strategies, which implicitly assume centred weights, are not effective for \acp{icnn}.
    By studying signal propagation through layers with non-negative weights, we are able to derive a principled weight initialisation for \acp{icnn}.
    Concretely, we generalise signal propagation theory by removing the assumption that weights are sampled from a centred distribution.
    In a set of experiments, we demonstrate that our principled initialisation effectively accelerates learning in \acp{icnn} and leads to better generalisation. 
    Moreover, we find that, in contrast to common belief, \acp{icnn} can be trained without skip-connections when initialised correctly.
    Finally, we apply \acp{icnn} to a real-world drug discovery task and show that they allow for more effective molecular latent space exploration.
\end{abstract}

\acresetall

\section{Introduction}
\label{sec:intro}

\begin{figure}
    \centering%
    \raisebox{10ex}{
        \begin{tabular}{l}
            \textsf{pre-activation} \\
            \textsf{distributions} \\[1.7em]
            \textsf{weight signs} \\
            {\color[HTML]{9ebeff} $\mathsf{{} < 0}$} or {\color[HTML]{d65244} $\mathsf{{} \geq 0}$} \\[1.7em]
            \textsf{feature} \\
            \textsf{correlations} \\[1.5em]
        \end{tabular}%
    }%
    \begin{subfigure}[t]{.27\textwidth}
        \centering
        \includegraphics[width=\linewidth]{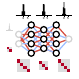}
        \caption{traditional}
        \label{fig:propagation:nn}
    \end{subfigure}%
    \hfill
    \begin{subfigure}[t]{.27\textwidth}
        \centering
        \includegraphics[width=\linewidth]{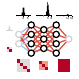}
        \caption{\acs{icnn}}
        \label{fig:propagation:icnn}
    \end{subfigure}%
    \hfill
    \begin{subfigure}[t]{.27\textwidth}
        \centering
        \includegraphics[width=\linewidth]{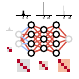}
        \caption{\acs{icnn} + init}
        \label{fig:propagation:icnn_init}
    \end{subfigure}
    \caption{
        Illustration of the effects due to good signal propagation in hidden layers.
        Blue and red connections depict negative and positive weights, respectively.
        The top row shows histograms of pre-activations in each hidden layer.
        The bottom row displays the feature correlation matrices for these layers.
        Small visualisations on the left depict the input distribution.
    }
    \label{fig:propagation}
\end{figure}

\textbf{Input-Convex Networks.}
\Acp{icnn} are networks for which each output neuron is convex with respect to the inputs.
The convexity is a result of using non-decreasing convex activation functions and weight matrices with non-negative entries \citep{amos17icnn}.
\Acp{icnn} were originally introduced in the context of energy modelling \citep{amos17icnn}.
Also in other contexts, \acp{icnn} have proven to be useful.
E.g.~\citet{sivaprasad21curious} show that \acp{icnn} can be used as regular classification models, 
\citet{makkuva20optimal} rely on the convexity to model optimal transport mappings,
and \citet{nesterov22learning} illustrate how \acp{icnn} can be used to learn invariances by simplifying the search for level sets --- i.e.~inputs for which the output prediction remains unchanged.
Despite their successful application for various tasks, convergence can be notably slow \citep[Fig.~1~(d)]{sivaprasad21curious}.
We hypothesize that this slow training is the result of poor initialisation of the positive weights in \acp{icnn}.
This poor initialisation leads to distribution shifts that make learning harder, as illustrated in Figure~\ref{fig:propagation}.
Therefore, we propose a principled initialisation strategy for 
layers with non-negative weight matrices \citep[cf.][]{chang20principled}.

\textbf{Importance of initialisation strategies.}
Initialisation strategies have enabled faster and more stable learning in deep networks \citep{lecun98efficient, glorot10understanding, he15delving}.
The goal of a good initialisation strategy is to produce similar statistics in every layer.
This can be done in the forward pass \citep{lecun98efficient, mishkin16lsuv, klambauer17self, chang20principled} or during back-propagation \citep{glorot10understanding, hoedt18characterising, defazio21beyond}.
It is also important to account for the effects due to non-linearities in the network \citep{saxe14exact, he15delving, klambauer17self, hoedt18characterising}.
However, there are no initialisation strategies for non-negative weight matrices, 
which are a key component of \acp{icnn} \citep{amos17icnn}.
We derive an initialisation strategy that accounts for the non-negative weights in \acp{icnn} by generalising the signal propagation principles that underlie modern initialisation strategies.

\textbf{Signal propagation.}
The derivation of initialisation strategies typically builds 
on the signal propagation framework introduced by \citet{neal95bayesian}.
This signal propagation theory has been used and expanded in various ways \citep{saxe14exact, poole16exponential, klambauer17self, martens21rapid}.
One critical assumption in this traditional signal propagation theory is that weights are sampled from a centred distribution, i.e.~with zero mean.
In \acp{icnn}, this is not possible because some weight matrices are constrained to be non-negative \citep{amos17icnn}.
Therefore, we generalise the traditional signal propagation theory to allow for non-centred distributions.

\textbf{Skip-connections in ICNNs.}
Architectures of \acp{icnn} typically include skip-connections \citep{amos17icnn, sivaprasad21curious, makkuva20optimal, nesterov22learning}.
The skip-connections in \acp{icnn} were introduced to increase their representational power \citep{amos17icnn}.
Although it is possible to study signal propagation with skip-connections \citep[e.g.][]{yang2017mean, brock21characterizing, hoedt22normalization}, they are typically built on top of existing results for plain networks.
Therefore, we limit our theoretical results to \acp{icnn} without skip-connections.
However, we find that \acp{icnn} without skip-connections can be successfully trained, indicating that skip-connections might not be necessary for representational power.
We show that with our initialisation, we are able to train an \ac{icnn} to the same performance as a non-convex baseline without skip-connections.
This confirms the hypothesis that good signal propagation can replace skip-connections \citep{martens21rapid, zhang22deep}.

\textbf{Contributions.}
Our contributions\footnote{Code for figures and experiments can be found at \url{https://github.com/ml-jku/convex-init}} can be summarised as follows:


\begin{itemize}
    \item We generalise signal propagation theory to include weights without zero mean (Section~\ref{sec:sigprop}).
    \item We derive a principled initialisation strategy for \acp{icnn} from our new theory  (Section~\ref{sec:icnn_init}).
    \item We empirically demonstrate the effectiveness of our initialisation strategy (Section~\ref{sec:experiments:init}).
    \item We apply \acp{icnn} in a real-world drug-discovery setting (Sections~\ref{sec:experiments:tox21} and~\ref{sec:experiments:levelsets}).
\end{itemize}



\section{Generalising Signal Propagation}
\label{sec:sigprop}


In this section, we revisit the traditional signal propagation theory \citep{neal95bayesian}, 
which assumes centred weights.
This provides us with a framework to study signal propagation in standard fully-connected networks.
We then expand this framework to enable the study of networks where weights do not have zero mean to derive a 
weight initialisation strategy for \acp{icnn} (Section~\ref{sec:icnn_init}).

\subsection{Background: Traditional Signal Propagation}
\label{sec:standard_sigprop}

When studying signal propagation in a network, the effect of a neural network layer on characteristics of the signal, such as its mean or variance, are investigated. 
The analysis of the vanishing gradient problem \citep{hochreiter91untersuchungen} can also be considered as signal propagation theory where the norm of the delta-errors is tracked throughout layers in backpropagation. 
Typically, the network is assumed to be a repetitive composition of the same layers (e.g.~fully-connected or~convolutional).
This allows to reduce the analysis of an entire network to a single layer and enables fixed-point analyses \citep[e.g.][]{klambauer17self, schoenholz17deep}.
In our work, we focus on fully-connected networks, $f : \reals^N \to \reals^M$, with some activation function $\phi : \reals \to \reals$ that is applied element-wise to vectors.
We study the propagation of pre-activations\footnote{a common alternative is to consider the (post-)activations \citep[see][]{hoedt18characterising}} \citep[cf.][]{saxe14exact, he15delving, poole16exponential} throughout the network:
\begin{equation}
    \label{eq:forward:pre-acts}
    \vec{s} = \mat{W} \phi(\vec{s}^{-}) + \vec{b}.
\end{equation}
Here, $\mat{W} \in \reals^{M \times N}$ and $\vec{b} \in \reals^M$ are the weight matrix and bias vector, respectively.
The notation $\vec{s}^{-}$ is used to indicate the pre-activations from the preceding layer, such that $\vec{s}^{-} \in \reals^N$.
 
The first two moments of the signal can be expressed in terms of the randomness arising from the parameters.
At initialisation time, the weight parameters are considered \ac{iid} random variables $w_{ij} \sim \mathcal{D}_w$.
The weight distribution is often assumed to be uniform or Gaussian, but the exact shape does not matter in wide networks \citep[Principle~2]{golikov22nongaussian}.
The bias parameters, on the other hand, are commonly ignored in the analysis, which is justified when $\forall i: b_i = 0$.
The pre-activations are also random variables due to the randomness of the parameters.
If we assume the weights to be centred, such that $\E[w_{ij}] = 0$, and $\Var[w_{ij}] = \sigma_w^2$, the first two moments of the pre-activations are given by
\begin{align}
    \label{eq:standard_sigprop:mom1}
    \E[s_i] &= \E\biggl[\sum_k w_{ik} \phi\bigl(s_k^{-}\bigr) + b_i\biggr] = 0 \\
    \label{eq:standard_sigprop:mom2}
    \E\bigl[s_i^2\bigr] &= \E\biggl[\Big(\sum_k w_{ik} \phi(s_k^{-}) + b_i\Big)^2\biggr] = N \sigma_w^2 \E\bigl[\phi(s_1^{-})^2\bigr].
\end{align}
Note that the variance is independent of the index, $i$, in the pre-activation vector and thus $\forall k : \E\bigl[\phi(s_k^{-})^2\bigr] = \E\bigl[\phi(s_1^{-})^2\bigr].$
Moreover, it can be shown (see Appendix~\ref{app:sigprop}) that $\forall i \neq j : \E[s_i s_j] = 0$, i.e. features within a pre-activation vector are uncorrelated in expectation.

Signal propagation theory has been used to derive initialisation and normalisation methods \citep{lecun98efficient, glorot10understanding, he15delving, klambauer17self}.
Initialisation methods often aim at having pre-activations with the same mean and variance in every layer of the network.
Because the mean is expected to be zero in every layer, the focus lies on keeping the variance of the pre-activations constant throughout the network, i.e.\ $\forall i, j : \E\big[s_i^2\big] = \E\big[{s_j^{-}}^2\big]= \sigma^2_*$.
Plugging this into Eq.~\eqref{eq:standard_sigprop:mom2}, we obtain the fixed-point equation
\begin{equation}
    \label{eq:standard_sigprop:fixpoint}
    \sigma^2_* = N \sigma_w^2 \operatorname{varprop}_\phi(\sigma^2_*).
\end{equation}
Here, $\operatorname{varprop}_\phi$ is a function that models the propagation of variance through the activation function,~$\phi$, assuming zero mean inputs.
E.g.~$\operatorname{varprop}_{\relu}(\sigma^2) = \frac{1}{2} \sigma^2$ \citep{he15delving} or $\operatorname{varprop}_{\tanh}(\sigma^2) \approx \sigma^2$ \citep{lecun98efficient, glorot10understanding}. 
We refer to Appendix~\ref{app:sigprop:kernels} for a more detailed discussion on propagation through activation functions.

This shows how modern initialisation methods are a solution to a fixed-point equation of the variance propagation.
Our goal is to apply the same principles to find an initialisation strategy for the non-negative weights in \acp{icnn}.
However, Eq.~\eqref{eq:standard_sigprop:fixpoint} heavily relies on the assumption that the weights are centred, i.e. $\mu_w = 0$.
This assumption is impossible to satisfy for the non-negative weights in \acp{icnn}, unless all weights are set to zero.
If $\mu_w \neq 0$, the mean of the pre-activations is no longer zero by default (as in Eq.~\ref{eq:standard_sigprop:mom1}).
Furthermore, also the propagation of variance and covariance is affected. 
Therefore, we extend the signal propagation theory to accurately describe the effects due to non-centred weights.

\subsection{Generalised Signal Propagation}
\label{sec:generalised_sigprop}

We generalise the traditional signal propagation theory by lifting a few assumptions from the traditional approach.
Most notably, the weights are no longer drawn from a zero-mean distribution, such that $\E[w_{ij}] = \mu_w \neq 0$.
Additionally, we include the effects due to bias parameters, which will give us extra options to control the signal propagation.
Similar to the weights, we assume bias parameters to be \ac{iid} samples from some distribution with mean $\E[b_i] = \mu_b$ and variance $\Var[b_i] = \sigma_b^2$.
By re-evaluating the expectations in Eq.~\eqref{eq:standard_sigprop:mom1} and~Eq.\eqref{eq:standard_sigprop:mom2}, we arrive at the following results:
\begin{align}
    \label{eq:sigprop:mom1}
    \E[s_i] &= N \mu_w \E[\phi(s_1^{-})] + \mu_b \\
    \label{eq:sigprop:mom2}
    \E[s_i s_j] &= \kronecker_{ij} \big(N \sigma_w^2 \E\big[\phi(s_1^{-})^2\big] + \sigma_b^2\big) + \mu_w^2 \sum_{k,k'} \Cov[\phi(s_k^{-}), \phi(s_{k'}^{-})] + \E[s_i] \E[s_j],
\end{align}
where $\kronecker_{ij}$ is the Kronecker delta.
We refer to Appendix~\ref{app:sigprop:generalised} for a full derivation.
Note that mean and variance are still independent of the index, $i$.
Therefore, we can continue to assume $\forall k : \E\bigl[\phi(s_k^{-})^n\bigr] = \E\bigl[\phi(s_1^{-})^n\bigr]$.

There are a few crucial differences between the traditional approach and our generalisation.
First, the expected value of the pre-activations is no longer zero by default.
This also means that the second moment, $\E\bigl[s_i^2\bigr],$ does not directly model the variance of the pre-activations in our setting.
Secondly, the propagation of the second moment has an additional term that incorporates effects due to the covariance structure of pre-activations in the previous layer.
Finally, the off-diagonal elements of the feature covariance matrix, $\Cov[s_i, s_j] = \E_{i \neq j}[s_i s_j] - \E[s_i] \E[s_j],$ are not necessarily zero and also depend on the variance.
This interaction between on- and off-diagonal elements in the feature covariance matrix makes it harder to ensure stable propagation.

Similar to Eq.~\eqref{eq:standard_sigprop:mom2}, Eq.~\eqref{eq:sigprop:mom2} depends on the variance propagation through the activation function. 
In addition, our generalised propagation theory involves the covariance propagation through the activation function.
The covariance propagation through the $\relu$ function has been studied in prior work \citep{cho2009kernel, daniely16toward} and can be specified as follows:
\begin{equation}
    \label{eq:sigprop:relu}
    \E[\relu(s_1) \relu(s_2)] = \frac{\sigma^2}{2 \pi} \big(\sqrt{1 - \rho^2} + \rho \arccos(-\rho)\big),
\end{equation}
with $(s_1, s_2) \sim \normaldist\bigg((0, 0), \Big(\begin{smallmatrix} 1 & \rho \\ \rho & 1 \end{smallmatrix}\Big) \sigma^2\bigg)$ and correlation $\rho$.
Note that this expression implicitly describes the propagation of the squared mean ($\rho = 0)$ and the second raw moment ($\rho = 1$).
We will assume $\phi = \relu$ for the remainder of our analysis.
A derivation for the propagation through the leaky $\relu$ activation function \citep{maas13lrelu} can be found in Appendix~\ref{app:sigprop:kernels}.

With our generalised propagation theory, the effects due to $\mu_w \neq 0$ become apparent.
As expected, the pre-activations no longer have zero mean by default.
Furthermore, the covariance plays a significant role in the signal propagation.
This shows that we also have to stabilise mean and covariance on top of the variance propagation to derive our principled initialisation.

\section{Principled Initialisation of ICNNs}
\label{sec:icnn_init}

With our generalised signal propagation theory, we will now derive a principled initialisation strategy for \acp{icnn}.
First, we set up fixed point equations for mean, variance and correlation using our generalised framework.
By solving these fixed point equations for $\mu_w, \sigma_w^2, \mu_b$ and $\sigma_b^2$ we obtain our principled initialisation.

\subsection{Background: Input-Convex Neural Networks}
\label{sec:icnn}

Neural networks for which all input-output mappings are convex, are called \acfp{icnn} \citep{amos17icnn}.
In general, neural networks are functions that are constructed by composing one or more simpler layers or components.
A function composition $f \circ g$ is convex if both $f$ and $g$ are convex, where $f$ additionally has to be non-decreasing in every output.
Therefore, \acp{icnn} are created by enforcing that every layer is \emph{convex} and \emph{non-decreasing} \citep{amos17icnn}.
A direct consequence is that only convex, non-decreasing activation functions can be used in \acp{icnn} --- e.g. $\softplus$ \citep{dugas01softplus}, $\relu$ \citep{nair10relu}, $\leakyrelu$ \citep{maas13lrelu}, $\elu$ \citep{clevert16elu}.

Fully-connected and convolutional layers are affine transformations, which are trivially convex.
To make these layers non-decreasing for building \acp{icnn}, their weights have to be non-negative \citep{amos17icnn}.
This can be done by replacing negative values with zero after every update.
The non-negativity constraint does not apply to bias parameters.
Another notable exception are direct connections from the input layer \citep{amos17icnn}.
This means that the first layer and any skip-connections from the input can have unconstrained weights.
\citet{amos17icnn} argue that these skip-connections are necessary for representational power.
We limit our analysis to networks without skip-connections.
Our experiments (see Section~\ref{sec:experiments}) 
show that \acp{icnn} with our proposed initialisation 
do not require skip-connections for efficient training.

\subsection{(Co-)Variance Fixed Points}
\label{sec:icnn_sigprop}



To set up the fixed-point equation for the second moments (Eq.~\ref{eq:sigprop:mom2}), we need to be able 
to propagate through the $\relu$ non-linearity.
However, the analytical results from Eq.~\eqref{eq:sigprop:relu} only hold if the pre-activations have zero mean.
Therefore, we look for a configuration that makes the mean (Eq.~\ref{eq:sigprop:mom1}) of the pre-activations zero.
Because we cannot enforce $\mu_w = 0$, we make use of the bias parameters to obtain centred pre-activations:
\begin{equation}
    \label{eq:bias:mean}
    \mu_b = -N \mu_w \E[\phi(s_1^{-})].
\end{equation}
This also allows us to refer to the second moments as the (co-)variance.
Note that we approximate the pre-activations with Gaussians
to use Eq.~\eqref{eq:sigprop:relu} at this point and for 
the remainder of the analysis.
Figure~\ref{fig:propagation:icnn_init} suggests that this Gaussian assumption generally does not hold in \acp{icnn} (see also Appendix~\ref{app:sigprop:kernels}).
Nevertheless, our experiments (see Section~\ref{sec:experiments}) suggest that the approximation is sufficiently good to derive an improved initialisation strategy for \acp{icnn}.
Furthermore, Figure~\ref{fig:propagation_bias:icnn_rbias} indicates that the use of random initial bias parameters (cf.~Appendix~\ref{app:experiments:rand_bias}) causes pre-activations to be (slightly) more Gaussian.

The propagation of the second moment described by Eq.~\eqref{eq:sigprop:mom2} consists of two parts.
The first part describes the off-diagonal entries, for which the dynamics are given by
\begin{equation*}
    \Cov_{i \neq j}[s_i, s_j] = \mu_w^2 \sum_{k, k'} \Cov[\phi(s_k^{-}), \phi(s_{k'}^{-})] = N \mu_w^2 \Big(\Var[\phi(s_1^{-})] + (N - 1) \Cov[\phi(s_1^{-}), \phi(s_2^{-})]\Big).
\end{equation*}
Here, we rewrite the sum assuming that the pre-activations are identically distributed.
This is possible because the covariance does not depend on the indices $i$ or $j$.
For further details, we refer to Appendix~\ref{app:sigprop:generalised}
The second part models the on-diagonal entries, i.e. the variance, which can be simplified in a similar way
\begin{equation*}
    \Var[s_i] = N \sigma_w^2 \E[\phi(s_1^{-})] + \sigma_b^2 + \Cov[s_1, s_2].
\end{equation*}
By plugging the $\relu$ moments from Eq.~\eqref{eq:sigprop:relu} into these results in Appendix~\ref{app:icnn_init:fixpoints}, 
we obtain the desired fixed-point equations in terms of variance and correlation:
\begin{align}
    \label{eq:icnn:corr}
    \rho_* &= \mu_w^2 \frac{1}{2 \pi} N \Big(\pi - N + (N - 1) \big(\sqrt{1 - \rho_*^2} + \rho_* \arccos(-\rho_*)\big)\Big) \\
    \label{eq:icnn:var}
    \sigma^2_* &= N \sigma_w^2 \frac{1}{2} \sigma^2_* + \sigma_b^2 + \rho_* \sigma^2_*.
\end{align}
with $\rho_* = \frac{1}{\sigma^2_*} \Cov[s_1^{-}, s_2^{-}]= \frac{1}{\sigma^2_*} \Cov_{i \neq j}[s_i, s_j]$ and $\sigma^2_* = \Var[s_1^{-}] = \Var[s_i].$

\subsection{Weight Distribution for ICNNs}
\label{sec:icnn_weightdist}

To obtain the distribution parameters for the initial weights, we solve the fixed point equations in Eq.~\eqref{eq:icnn:corr} and~Eq.\eqref{eq:icnn:var} for $\sigma_b^2$, $\sigma_w^2$ and $\mu_w^2$.
Because this system is over-parameterised, we choose to set $\sigma_b^2 = 0$.
An analysis for $\sigma_b^2 > 0$ can be found in Appendix~\ref{app:experiments:rand_bias}.
This leads to the following initialisation parameters for \acp{icnn}:
\begin{align}
    \label{eq:weightdist:var}
    \sigma_w^2 &= \frac{2}{N} (1 - \rho_*) \\
    \label{eq:weightdist:mean}
    \mu_w^2 &= \frac{2 \pi}{N} \rho_* \Big(\pi - N + (N - 1) \big(\sqrt{1 - \rho_*^2} + \rho_* \arccos(-\rho_*)\big)\Big)^{-1}.
\end{align}
Note that these solutions only depend on the correlation, $\rho_*$, and not on the variance, $\sigma^2_*$.
For a stability analysis of these fixed points, we refer to Appendix~\ref{app:icnn_init:fixpoint_stability}.

Although our initialisation (Eq.~\ref{eq:weightdist:mean} and~\ref{eq:weightdist:var}) is entirely 
defined by the correlation between features, $\rho = \frac{\Cov[s_1, s_2]}{\Var[s_1]}$, 
not all solutions are admissible.
For example, uncorrelated features ($\rho = 0$) are only possible if $\mu_w = 0$.
This means that features in \acp{icnn} must have non-zero correlation.
On the other hand, perfect correlation ($\rho = 1$) would mean that all features point in the same direction, which is not desirable.
Therefore, we choose $\rho_* = \frac{1}{2}$ as a compromise to obtain the following initialisation parameters for our experiments in Section~\ref{sec:experiments}:
\begin{align*}
    \mu_w &= \sqrt{\frac{6 \pi}{N \big(6 (\pi - 1) + (N - 1) (3 \sqrt{3} + 2 \pi - 6)\big)}} &
    \sigma_w^2 &= \frac{1}{N} \\
    \mu_b &= \sqrt{\frac{3 N}{6 (\pi - 1) + (N - 1) (3 \sqrt{3} + 2 \pi - 6)}} &
    \sigma_b^2 &= 0.
\end{align*}
We refer to appendix~\ref{app:experiments:ablations} for a discussion on different choices for $\rho_*$.

For the first layer, which can also have negative weights, we use \citet{lecun98efficient} initialisation.
For the non-negative weights in an \acp{icnn}, we need to sample from a distribution with non-negative support.
The lower bound of a uniform distribution with our suggested mean and variance is negative for any $N > 1$ and thus not useful.
Also, sampling from a Gaussian distribution is not practical because its support includes negative values.
Eventually, we propose to sample from a log-normal distribution with parameters
\begin{align*}
    \tilde{\mu_w} &= \ln(\mu_w^2) - \frac{1}{2} \ln(\sigma_w^2 + \mu_w^2) &
    \tilde{\sigma_w}^2 &= \ln(\sigma_w^2 + \mu_w^2) - \ln(\mu_w^2).
\end{align*}
This ensures that sampled weights are non-negative and have the desired mean and variance.
Note that other positive distributions with sufficient degrees of freedom should also work.

One advantage of the log-normal distribution is that sampling is simple and efficient.
It suffices to sample $\tilde{w}_{ij} \sim \normaldist(\tilde{\mu_w}, \tilde{\sigma_w}^2)$ to compute $w_{ij} = \exp\bigl(\tilde{w}_{ij}\bigr)$.
In the original \ac{icnn} \citep{amos17icnn}, the exponential function is only used to make the initial weights positive.
However, if weights are re-parameterised using the exponential function, initial weights can be directly sampled from a Gaussian distribution.
This setting is studied in Appendix~\ref{app:experiments:reparam}.

\section{Related Work}
\label{sec:related}


\emph{Input-convex neural networks.} \acf{icnn} were originally designed in the context of energy models \citep{amos17icnn}.
We focus on the fully convex \acp{icnn} variant and use gradient descent for optimisation instead of the proposed bundle-entropy method.
In their implementation, \citet{amos17icnn} use projection methods to keep weights positive,
i.e.\ negative values are set to zero after every update.
However, also other projection methods can be used to keep the weights positive \citep[e.g.][]{sivaprasad21curious}.
Instead of projecting the weights onto the non-negative reals after every update, it is also possible to use a reparameterisation of the weights.
E.g. \citet{nesterov22learning} square the weights in the forward pass.
Note that a reparameterisation can have an effect on the learning dynamics because it is an inherent part of the forward pass.
Another alternative is to use regularisation to impose a \emph{soft} non-negativity constraint on the weights \citep{makkuva20optimal}.
Similar to \citep{sivaprasad21curious}, we directly train an \ac{icnn}, but we 
do not observe the same generalisation benefits.
On the other hand, \citet{sankaranarayanan22cdinn} point out that 
restricting the network to be convex decreases their capacity.
They propose to use the difference of two \acp{icnn} to allow modelling non-convex functions \citep[c.f.][]{yuille01cccp}.
However, we did not find the convexity restriction to cause any problems in our experiments.
\citet{nesterov22learning} use \acp{icnn} to simplify the computation of level sets to learn invariances.
We adopt this experiment setting to highlight the potential of \acp{icnn}.
%
\emph{Optimization.} While training the parameters of an \ac{icnn} is unrelated to convex optimisation,
\citet{bengio05convex} showed how training 
a neural network together with the number of hidden 
neurons can be seen as a convex optimisation problem.
This work was continued by \citet{bach17breaking}, 
who established a connection between this convex optimisation and automatic feature selection.
\emph{Signal propagation theory.} The study of signal propagation is concerned with tracking statistics of the data throughout multiple layers of the network.
In his thesis, \citet{neal95bayesian} computed how the 
first two moments propagate through a two-layer network to study networks as Gaussian processes.
This is similar to how modern initialisation strategies have been derived \citep[e.g.][]{lecun98efficient,he15delving}.
A similar approach has been taken to incorporate the propagation of gradients for initialisation \citep{glorot10understanding}.
The idea to explicitly account for the effects of activation functions can be attributed to \citep{saxe14exact}.
All of these analyses assume that weights are initialised from a zero-mean distribution, which is not possible in \acp{icnn}.
\citet{mishkin16lsuv} empirically compute the variance for the weights.
This approach is more robust to deviations, but nevertheless requires adaptations to account for weights with non-zero mean.
%
%
There is also a significant body of work that studies signal propagation in terms of mean field theory \citep{poole16exponential}.
In these works, the correlation between samples is included in the analysis \citep{poole16exponential}.
In our work, we find that the correlation between features plays an important role.
\citet{schoenholz17deep} applied the mean field analysis to the backward dynamics and introduced the concept of \emph{depth scales} to quantify how deep signals can propagate.
The mean field theory was also applied to convolutional networks to derive the delta-orthogonal initialisation \citep{xiao18dynamical}.
We refer to \citep{martens21rapid} for 
a general overview of signal propagation.

\section{Experiments}
\label{sec:experiments}

We evaluate our principled initialisation strategy by training \acp{icnn} on three sets of experiments. 
In our first experiments, we investigate the effect of initialisation on learning dynamics and generalisation in \acp{icnn} on multiple permuted image datasets.
We also include non-convex networks in these experiments to illustrate that \acp{icnn} with our principled initialisation can be trained as well as regular networks.
However, we would like to stress that non-convex networks are expected to outperform \acp{icnn} because they are not constrained to convex decision boundaries.
Moreover, we assess whether \acp{icnn} using our initialisation are able to match the performance of regular fully-connected networks in toxicity prediction without the need for skip-connections.
Finally, we explore \acp{icnn} as a tool for more controlled latent space exploration of a molecule auto-encoder system.
This illustrates a setting where regular networks can no longer be used. 
Details on the computing budget are provided in the Appendix.

\subsection{Computer vision benchmarking datasets}
\label{sec:experiments:init}

As an illustrative first set of experiments, we use computer vision benchmarks to assess the effects of our principled initialisation strategy.
Concretely, we trained fully-connected \acp{icnn} on {MNIST} \citep{bottou94mnist}, {CIFAR10} and {CIFAR100} \citep{krizhevsky09cifar}, 
to which we refer as permuted image benchmarks \citep[cf.][]{goodfellow14empirical}.

In our comparison, we consider four different types of methods:
(1) a classical non-convex network without skip-connections and default initialisation \citep[cf.][]{he15delving}.
(2) an \ac{icnn}, ``\ac{icnn}'',  without skip connections and default initialisation 
(3) an \ac{icnn}, ``\ac{icnn} + skip'',  with skip connections and default initialisation
(4) an \ac{icnn}, ``\ac{icnn} + init'',  without skip connections and our proposed initialisation.
All networks use $\relu$ activation functions.
The non-negativity constraint in \acp{icnn} is implemented by setting negative weights to zero after every update \citep[cf.][]{amos17icnn}. 
Note that skip-connections in \acp{icnn} introduce 
additional parameters to allow for additional 
representational power \citep[cf.][]{amos17icnn}.

\begin{figure}
    \centering
    \includegraphics[width=\linewidth]{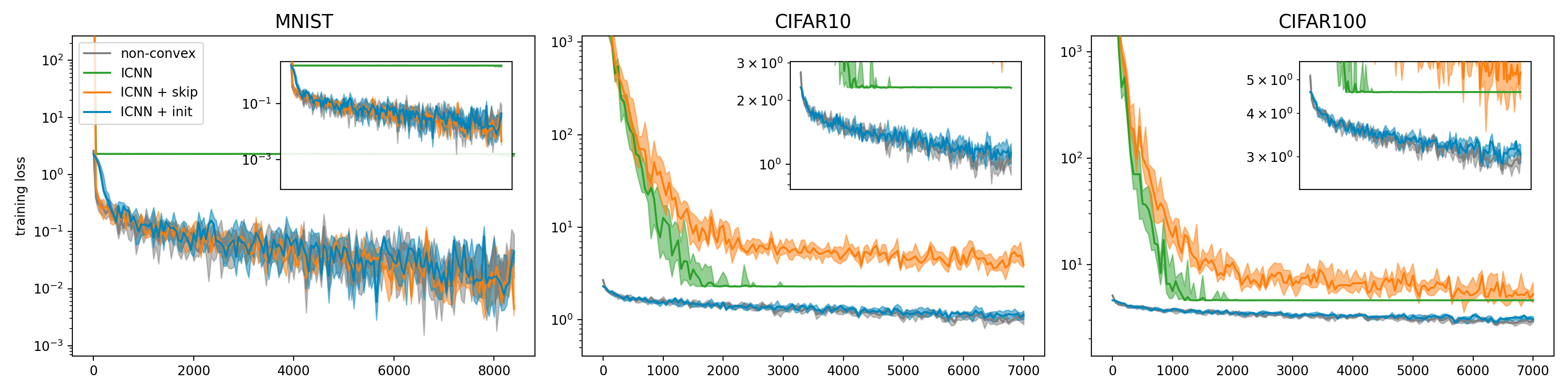}
    \caption{
        Training loss curves of \ac{icnn} variants with the same architecture on the {MNIST}, {CIFAR10} and {CIFAR100} datasets.
        ''\ac{icnn}`` input-convex network with default initialisation.
        ''\ac{icnn} + skip``: same settings, but with 
        skip connections. ''\ac{icnn} + init``
        our principled initialisation for \acp{icnn} w/o skip connections.
        ''non-convex``: a regular non-convex network.
        Each curve represents the median performance over ten runs and shaded regions indicate the inter-quartile range.
        The inset figures provide a view of the loss curves zoomed in. 
        Note that \ac{icnn} losses do decrease before the plateau.
    }
    \label{fig:learn_curves}
\end{figure}

\textbf{Training dynamics.}
\emph{Settings.} 
In the first experiment, we analyze the training dynamics during the first ten epochs of training.
Similar to \citet{chang20principled}, we fixed the number of neurons in each layer to the input size, and the number of hidden layers to five.
We refer to Appendix~\ref{app:experiments:init} for results with different depths.
The learning rate for the Adam optimiser was obtained by manually tuning on the non-convex baseline.
\emph{Results.} 
We found that \acp{icnn} with our principled initialisation exhibit similar learning behaviour as the non-convex baseline, 
while \acp{icnn} without our initialisation strategy can not decrease the loss to the level of the baseline on {CIFAR10} and {CIFAR100}.
Figure~\ref{fig:learn_curves} shows the learning curves for our different methods on three permuted image benchmarks.



\begin{figure}
    \centering
    \includegraphics[width=.49\linewidth]{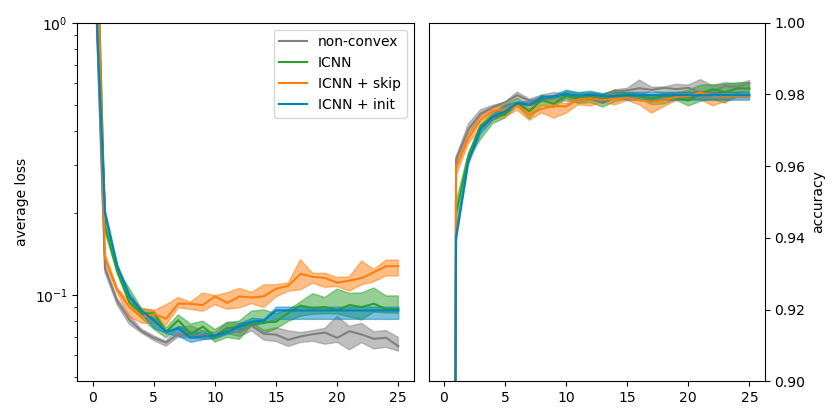}%
    \includegraphics[width=.49\linewidth]{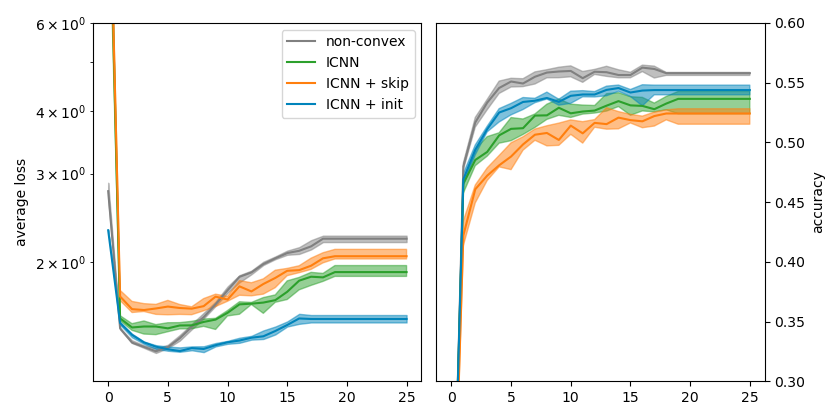}
    \caption{
        Test set metrics of 
       compared methods on the (a) {MNIST} and (b) {CIFAR10} datasets. 
        Each curve displays the median performance over ten runs.
        Shaded regions represent the inter-quartile range over the ten runs. On MNIST, all methods can be successfully trained and exhibit similar performance. On CIFAR 10, \acp{icnn} with our proposed initialisation outperform other \acp{icnn} variants. 
    }
    \label{fig:fine_tuned}
\end{figure}

\textbf{Generalisation.} \emph{Settings.} 
In our next experiment, the effects of our principled initialisation on the generalisation capabilities of \acp{icnn} is investigated.
To this end, we train our three \ac{icnn} variants and the non-convex baseline on the permuted image benchmarks again, but focus on the test performance this time.
To this end we perform a grid-search to find the hyper-parameters that attain the best accuracy after 25 epochs of training on a random validation split, for each of the four compared methods.
The grid consists of multiple architectures with at 
least one hidden layer, learning rate for Adam, $L_2$-regularisation and whitening pre-processing transforms (details in Appendix~\ref{app:experiments:init}).
\emph{Results.} On the {MNIST} dataset, all variants reach similar accuracy values, which we attribute to the simplicity of the prediction problem. 
However, on {CIFAR10} the initialisation strategy
leads to better generalisation already in early epochs and --- compared to the respective \acp{icnn} variants without our initialisation --- improves generalisation overall (see Figure~\ref{fig:fine_tuned}). 
The \ac{icnn} with our initialisation almost matches the accuracy values of non-convex nets.

\subsection{Toxicity Prediction}
\label{sec:experiments:tox21}

To test \acp{icnn} in a real-world setting, in a different application domain, we consider the binary multi-task problem of toxicity prediction in drug discovery.
More specifically, we train \acp{icnn} on the {Tox21} challenge data \citep{huang16tox21,mayr2016deeptox,klambauer17self}.
The input data consists of so-called {SMILES} representations of small molecules.
The target variables are binary labels for twelve different measurements or assays indicating whether a molecule induces a particular toxic effect on human cells.
During pre-processing, {SMILES} strings were converted to \acp{cddd} \citep{winter19cddd}, which are 512-dimensional numerical representations of the molecules.
The choice of these particular descriptors is motivated by \acp{cddd} properties of being decodeable into valid molecules (see Section~\ref{sec:experiments:levelsets}). 
Hyper-parameters were selected by a manual search on the non-convex baseline.
We ended up using a fully-connected network with two hidden layers of 128 neurons and $\relu$ activations.
The network was regularised with fifty percent dropout after every hidden layer, as well as seventy percent dropout of the inputs.
The results in Table~\ref{tab:tox21results} show that our initialisation significantly outperforms \acp{icnn} with standard initialization ($p$-value 2.6e-13, binomial test) and \acp{icnn} with skip connections ($p$-value 5.5e-14).
Furthermore, results are close to the performance of traditional, non-convex networks.

\begin{table}
    \centering
    \caption{
        Area under the ROC curve on the test set for each of the twelve tasks in the {Tox21} data.
        Each value represents the median performance over 10 runs and error bars is the maximum distance to the boundary of the interval defined by the (0.05, 0.95) quantiles. Of all variants of \acp{icnn}, the variant 
        with our proposed initialisation, ``ICNN+init (ours)'', performs best matching almost the predictive quality of
        traditional non-convex neural networks. 
    }
    \label{tab:tox21results}
    \resizebox{\columnwidth}{!}{
    \begin{tabular}{lccccccc}
         & NR.AhR & NR.AR & NR.AR.LBD & NR.Aromatase & NR.ER & NR.ER.LBD \\
         \toprule
        \ac{icnn}+init (ours) & $\mathbf{91.29} \pm 0.58\%$ & $\mathbf{81.84} \pm 2.43\%$ & $\mathit{81.36} \pm 3.58\%$ & $\mathit{82.40} \pm 1.09\%$ & $\mathbf{78.37} \pm 0.80\%$ & $77.85 \pm 0.90\%$& \\

        \ac{icnn}  & $90.56 \pm 0.63\%$ & $\mathit{81.28} \pm 4.82\%$ & $78.14 \pm 3.15\%$ & $80.46 \pm 1.74\%$  & $77.30 \pm 0.78\%$ & $\mathit{78.15} \pm 1.02\%$ &\\
        \ac{icnn}+skip & $89.83 \pm 0.21\%$ & $68.12 \pm 1.74\%$ & $74.17 \pm 2.20\%$ & $78.95 \pm 0.45\%$ & $76.95 \pm 0.48\%$ & $\mathbf{81.92} \pm 1.16\%$ & \\
        \midrule
        non-convex & $\mathit{91.01} \pm 0.65\%$ & $79.21 \pm 1.73\%$ & $\mathbf{86.19} \pm 2.50\%$ & $\mathbf{83.63} \pm 0.34\%$ & $\mathit{77.88} \pm 1.13\%$ & $74.32 \pm 1.59\%$ & \vspace{0mm} \\ \midrule 
         & SR.ATAD5 & SR.HSE & SR.MMP & SR.p53 & NR.PPAR$\gamma$ & SR.ARE  & AVG \\
        \toprule
        \ac{icnn}+init (ours)& $77.01 \pm 1.67\%$ & $\mathit{80.05} \pm 1.36\%$ & $\mathit{93.69} \pm 0.27\%$ & $81.10 \pm 0.46\%$  & $77.46 \pm 2.50\%$ & $76.80 \pm 0.33\%$ & $\mathit{80.57} \pm 11.84\%$ \\ 
        \ac{icnn} & $74.25 \pm 3.26\%$ & $78.64 \pm 1.42\%$ & $93.19 \pm 0.59\%$ & $\mathit{81.32} \pm 0.43\%$ & $75.00 \pm 1.73\%$ & $76.11 \pm 0.82\%$ & $78.33 \pm 13.42\%$\\ 
        \ac{icnn}+skip & $\mathit{75.93} \pm 1.26\%$ & $78.23 \pm 0.50\%$ & $75.40 \pm 0.88\%$ & $78.25 \pm 0.93\%$ & $\mathbf{92.09} \pm 0.41\%$ & $\mathbf{79.32} \pm 0.97\%$ & $78.29 \pm 12.56\%$ \\ \midrule
        non-convex & $\mathbf{78.93} \pm 1.68\%$ & $\mathbf{80.12} \pm 2.57\%$ & $\mathbf{93.99} \pm 0.39\%$ & $\mathbf{82.17} \pm 0.96\%$ & $\mathit{82.70} \pm 1.21\%$ & $\mathit{78.30} \pm 0.90\%$ & $\mathbf{81.31} \pm 11.03\%$ \\
        \bottomrule
    \end{tabular}
    }
\end{table}


\subsection{Latent space exploration of molecular spaces}
\label{sec:experiments:levelsets}

\begin{figure}
    \centering
    \includegraphics[width=\linewidth]{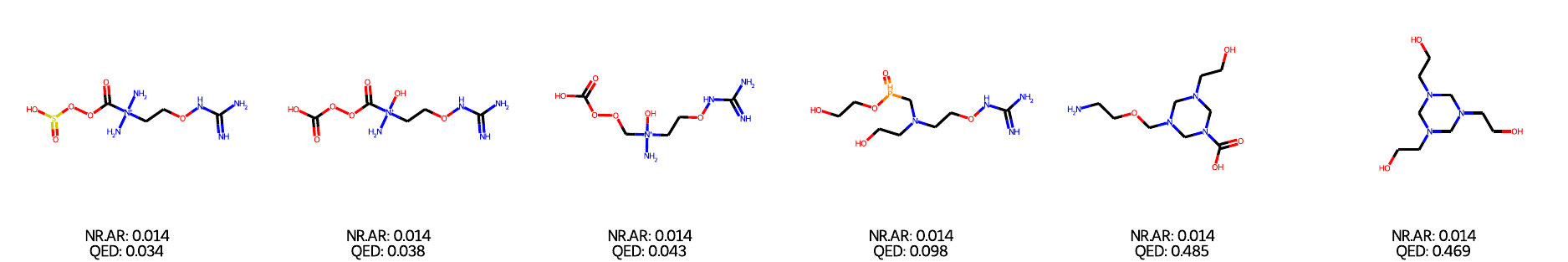}
    \includegraphics[width=\linewidth]{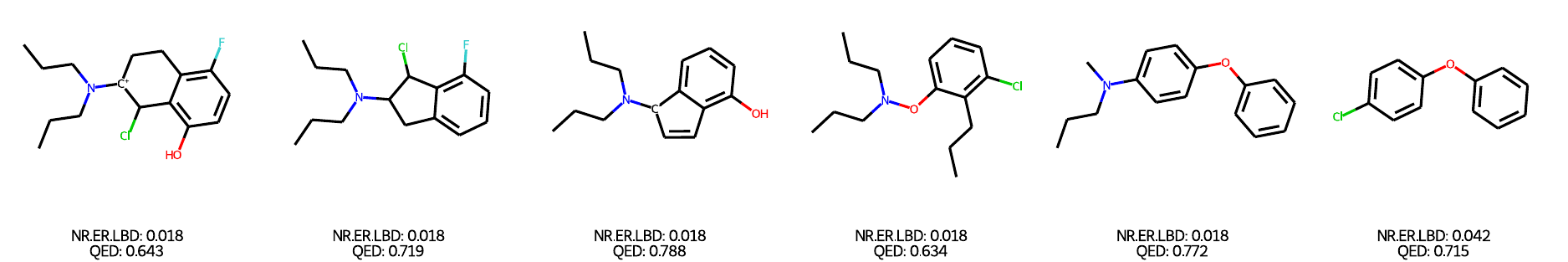}
    \caption{Example of two level set trajectories of a {Tox21} model. The leftmost molecule represents
    the starting molecule with low toxicity 
    (top: ``NR.AR'', bottom: ``NR.ER.LBD''). The rightmost molecule represents the target molecule. 
    All intermediary molecules are samples on the level set between the starting molecule
    and the target molecule and exhibit different drug-likeness (``QED''). In this way, one predicted molecular property can be kept fixed  while another property is optimized.}
    \label{fig:levelsets}
\end{figure}

In this experiment, we exploit the intrinsic property of \acp{icnn} that level sets, i.e.~the set of all inputs that map to the same output, can be parameterised \citep{nesterov22learning}.
These level sets provide an opportunity in drug discovery to keep one molecular property, such as low toxicity, fixed while optimizing another property, such as drug-likeness.
To demonstrate this application, we randomly chose reference molecules with low toxicity and followed the according level set in the direction of another randomly chosen target molecules. 
Since the input space of the \ac{icnn} that predicts toxicity (see above) possesses a decoder model from \citep{winter19cddd}, the numeric representations of the molecules on the level set can be decoded into molecular structures. 
The obtained trajectories in Figure~\ref{fig:levelsets} demonstrate that the predicted toxicity of the molecules remain constant, while different QED scores, which measure drug-likeness, are obtained by traversing the level set from the reference molecule to the target molecule.


\section{Conclusion and Discussion}
We demonstrated how signal propagation theory can be generalised to include weights without zero mean.
By controlling the propagation of correlation through the network, 
we derived an initialisation strategy that promotes a stable propagation of signals.
We empirically verified that this steady propagation leads to improved learning.
Finally, we showed that \acp{icnn} can be trained 
to a comparable level as regular networks on the 
{Tox21} data using \acp{cddd}.

Our initialisation approximates the distribution of
pre-activations with Gaussians, 
which falls short of agreement with the observed data, 
but has been sufficient to improve the initialisation of \acp{icnn}.
A possible way forward is to use the central limit theorem 
for weakly dependent variables.
In Appendix~\ref{app:experiments:rand_bias}, we also observed that random bias initialisation leads to pre-activations with a more Gaussian distribution.
Appendix~\ref{app:sigprop:conv} also includes an analysis for convolutional layers, 
but proficient empirical results with input-convex convolutional networks are still to be made.
We conjecture that this is due to the more 
complicated covariance structure in convolutional layers.
We also included a signal propagation analysis for back-propagation in Appendix~\ref{app:sigprop:backprop}.
However, incorporating insights from this analysis into an initialisation method is also left for future work.
We envision that our initialisation strategy will make it easier 
to incorporate \acp{icnn} or other networks with non-negative weights.



\begin{ack}
We thank Sepp Hochreiter for pointing us in the direction of input-convex networks and their interesting properties.
Special thanks go to Niklas Schmidinger for his help with experiments on convolutional networks.

The ELLIS Unit Linz, the LIT AI Lab, the Institute for Machine Learning, are supported by the Federal State Upper Austria. We thank the projects AI-MOTION (LIT-2018-6-YOU-212), DeepFlood (LIT-2019-8-YOU-213), Medical Cognitive Computing Center (MC3), INCONTROL-RL (FFG-881064), PRIMAL (FFG-873979), S3AI (FFG-872172), DL for GranularFlow (FFG-871302), EPILEPSIA (FFG-892171), AIRI FG 9-N (FWF-36284, FWF-36235), AI4GreenHeatingGrids(FFG- 899943), INTEGRATE (FFG-892418), ELISE (H2020-ICT-2019-3 ID: 951847), Stars4Waters (HORIZON-CL6-2021-CLIMATE-01-01). We thank Audi.JKU Deep Learning Center, TGW LOGISTICS GROUP GMBH, Silicon Austria Labs (SAL), FILL Gesellschaft mbH, Anyline GmbH, Google, ZF Friedrichshafen AG, Robert Bosch GmbH, UCB Biopharma SRL, Merck Healthcare KGaA, Verbund AG, GLS (Univ. Waterloo) Software Competence Center Hagenberg GmbH, T\"{U}V Austria, Frauscher Sensonic, TRUMPF and the NVIDIA Corporation.
\end{ack}

\bibliographystyle{apalike}
\bibliography{references}


\clearpage
\appendix

\resumetocwriting
\renewcommand*\contentsname{Contents of the Appendix}
\tableofcontents

\newpage
\section{Generalisation of Signal Propagation}
\label{app:sigprop}

This section provides the
derivations for the generalised signal propagation.

\subsection{Traditional Signal Propagation}
\label{app:sigprop:traditional}

One of the key assumptions in traditional signal propagation is that the weights are drawn from a zero-mean distribution with some variance $\sigma_w^2$.
E.g. $w \sim \normaldist(0, \sigma_w^2)$ or $w \sim \uniformdist\Bigr(-\sqrt{3\sigma_w^2}, \sqrt{3 \sigma_w^2}\Bigl)$ are typical distributions for sampling initial weights.
Also, bias parameters are typically assumed to be initialised with zeros, such that mean and variance are $\mu_b = 0$ and $\sigma_b^2 = 0$, respectively.
Since the initial weights are drawn from \ac{iid} samples, the first two (raw) moments of the pre-activations from eq.~\eqref{eq:forward:pre-acts} be directly computed as follows:
\begin{align}
    \nonumber
    \E[s_i] &= \E\Bigl[b_i + \sum_k w_{ik} \phi(s_k^{-})\Bigr] \\ 
    \nonumber
    &=\E[b_i] + \sum_k \E[w_{ik}] \E[\phi(s_k^{-})] \\
    \label{eq:standard_sigprop:mom1_raw}
    &= \mu_b + \sum_k \mu_w \E[\phi(s_k^{-})] = 0 \\ 
    \nonumber
    \E\bigl[s_i^2\bigr] &= \E\biggl[\Big(b_i + \sum_k w_{ik} \phi(s_k^{-})\Big)^2\biggr] \\ 
    \nonumber
    &= \E\bigl[b_i^2\bigr] + 2 \E[b_i] \sum_k \E[w_{ik}] \E[\phi(s_k^{-})] + \E\biggl[\Big(\sum_k w_{ik} \phi(s_k^{-})\Big)^2\biggr] \\ 
    \nonumber
    &= (\sigma_b^2 + \mu_b^2) + 2 \mu_b \sum_k \mu_w \E[\phi(s_k^{-})] + \E\Bigl[\sum_{k,k'} w_{ik} w_{ik'} \phi(s_k^{-}) \phi(s_{k'}^{-})\Bigr] \\ \nonumber
    &= \E\Bigl[\sum_{k,k'} w_{ik} w_{ik'} \phi(s_k^{-}) \phi(s_{k'}^{-})\Bigr] \\ \nonumber
    &= \sum_k \E\bigl[w_{ik}^2\bigr] \E\bigl[\phi(s_k^{-})^2\bigr] + \sum_{k,k'\neq k} \E[w_{ik}] \E[w_{ik'}] \E[\phi(s_k^{-}) \phi(s_{k'}^{-})] \\ 
    \nonumber
    &= \sum_k (\sigma_w^2 + \mu_w^2) \E\bigl[\phi(s_k^{-})^2\bigr] + \sum_{k,k' \neq k} \mu_w^2 \E[\phi(s_k^{-}) \phi(s_{k'}^{-})] \\
    \label{eq:standard_sigprop:mom2_raw}
    &= \underbrace{\sigma_w^2 \sum_{k} \E\bigl[\phi(s_k^{-})^2\bigr]}_{\Var[s_i]}.
\end{align}

If we additionally assume that the pre-activations from the previous layer are identically distributed, such that
\begin{equation}
    \label{eq:varsum:independent}
    \forall k : \E\bigl[\phi(s_k^{-})^2\bigr] = \E\bigl[\phi(s_1^{-})^2\bigr],
\end{equation}
we retrieve the traditional signal propagation formulas (equations~\ref{eq:standard_sigprop:mom1} and~\ref{eq:standard_sigprop:mom2}):
\begin{align*}
    \E[s_i] &= 0 \\
    \E\bigl[s_i^2\bigr] &= N \sigma_w^2 \E\bigl[\phi(s_1^{-})^2\bigr],
\end{align*}
where $N$ is the number of incoming connections, such that $\vec{s}^{-} \in \reals^N$.

For the first layer, the recursion formulas in this section can obviously not be used.
The moments of the pre-activations in the first layer can be computed using a very similar derivation, however:
\begin{align*}
    \E[s_i] &= \E\Bigl[b_i + \sum_k w_{ik} x_k\Bigr] &
    \E\bigl[s_i^2\bigr] &= \E\biggl[\Big(b_i + \sum_k w_{ik} x_k\Big)^2\biggr] \\
    &= \mu_b + \mu_w \sum_k \E[x_k] = 0 &
    &= \sigma_w^2 \sum_{k} \E\bigl[x_k^2\bigr].
\end{align*}
There are two possible ways to interpret the inputs.
If we wish to treat inputs as random variables, we would also need $x_k$ to be identically distributed to obtain
\begin{align*}
    \E[s_i] &= 0 &
    \E\bigl[s_i^2\bigr] &= N \sigma_w^2 \E\bigl[x_1^2]
\end{align*}
Alternatively, we can just use the input data as constants, such that $\E\bigl[x_k^2\bigr] = x_k^2$.
In this scenario, the pre-activation moments are
\begin{align*}
    \E[s_i] &= 0 &
    \E\bigl[s_i^2\bigr] &= \sigma_w^2 \norm{\vec{x}}.
\end{align*}

\subsection{Generalised Signal Propagation}
\label{app:sigprop:generalised}

In \acp{icnn}, the zero-mean assumption on the weights does not hold.
Therefore, we apply the signal propagation analysis assuming that weights have mean $\mu_w$ and variance $\sigma_w^2$.
Furthermore, we also account for non-zero bias initialisation by not making any assumptions on the mean and variance of the bias parameters.
In this setting, the first two (raw) moments of the pre-activations from eq.~\eqref{eq:forward:pre-acts} are:
\begin{align}
    \nonumber
    \E[s_i] &= \E\Bigl[b_i + \sum_k w_{ik} \phi(s_k^{-})\Bigr] \\
    \label{eq:sigprop:mom1_raw}
    &= \mu_b + \mu_w \sum_k \E[\phi(s_k^{-})] \\
    \nonumber
    \E\bigl[s_i^2\bigr] &= \E\biggl[\Big(b_i + \sum_k w_{ik} \phi(s_k^{-})\Big)^2\biggr] \\
    \nonumber
    &= \sigma_b^2 + \mu_b^2 + 2 \mu_b \mu_w \sum_k \E[\phi(s_k^{-})] + \E\Bigl[\sum_{k,k'} w_{ik} w_{ik'} \phi(s_k^{-}) \phi(s_{k'}^{-})\Bigr] \\
    \nonumber
    &= \sigma_b^2 + \mu_b^2 + 2 \mu_b \mu_w \sum_k \E[\phi(s_k^{-})] + (\sigma_w^2 + \mu_w^2) \sum_k \E\bigl[\phi(s_k^{-})^2\bigr] + \mu_w^2 \sum_{k,k' \neq k} \E[\phi(s_k^{-}) \phi(s_{k'}^{-})] \\
    \nonumber
    &= \sigma_b^2 + \mu_b^2 + 2 \mu_b \mu_w \sum_k \E[\phi(s_k^{-})] + \sigma_w^2 \sum_k \E\bigl[\phi(s_k^{-})^2\bigr] + \mu_w^2 \sum_{k,k'} \E[\phi(s_k^{-}) \phi(s_{k'}^{-})] \\
    \nonumber
    &= \sigma_b^2 + \sigma_w^2 \sum_k \E\bigl[\phi(s_k^{-})^2\bigr] + \mu_w^2 \sum_{k,k'} \E[\phi(s_k^{-}) \phi(s_{k'}^{-})] \\
    \nonumber
    &\qquad - \mu_w^2 \sum_{k,k'} \E[\phi(s_k^{-})] \E[\phi(s_{k'}^{-})] + \Big(\mu_b + \mu_w \sum_{k} \E[\phi(s_k^{-})]\Big)^2 \\
    \label{eq:sigprop:mom2_raw}
    &= \underbrace{\sigma_b^2 + \sigma_w^2 \sum_k \E\bigl[\phi(s_k^{-})^2\bigr] + \mu_w^2 \sum_{k,k'} \Cov[\phi(s_k^{-}), \phi(s_{k'}^{-})]}_{\Var[s_i]} + \E[s_i]^2,
\end{align}
where
\begin{equation*}
    \Cov[\phi(s_k^{-}), \phi(s_{k'}^{-})] = \E[\phi(s_k^{-})\phi(s_{k'}^{-})] - \E[\phi(s_k^{-})] \E[\phi(s_{k'}^{-})]
\end{equation*}
is the covariance.

Unlike in the traditional setting in section~\ref{app:sigprop:traditional}, the pre-activations are not uncorrelated.
This means that we also need to include the propagation of covariance in our generalised signal propagation theory.
We do this by considering the second mixed moment of the pre-activations:
\begin{align}
    \nonumber
    \E_{i \neq j}[s_i s_j] &= \E\Bigl[\Big(b_i + \sum_k w_{ik} \phi(s_k^{-})\Big) \Big(b_j + \sum_k w_{jk} \phi(s_k^{-})\Big)\Bigr] \\
    \nonumber
    &= \E[b_i b_j] + 2 \E\Bigl[b_i \sum_k w_{jk} \phi(s_k^{-})\Bigr] + \E\Bigl[\sum_{k, k'} w_{ik} w_{jk'} \phi(s_k^{-}) \phi(s_{k'}^{-})\Bigr] \\
    \nonumber
    &= \mu_b^2 + 2 \mu_b \mu_w \sum_k \E[\phi(s_k^{-})] + \mu_w^2 \sum_{k, k'} \E[\phi(s_k^{-}) \phi(s_{k'}^{-})] \\
    \nonumber
    &= \Big(\mu_b + \mu_w \sum_k \E[\phi(s_k^{-})]\Big)^2 + \mu_w^2 \sum_{k, k'} \Cov[\phi(s_k^{-}), \phi(s_{k'}^{-})] \\
    \label{eq:sigprop:mom_mix_raw}
    &= \underbrace{\mu_w^2 \sum_{k, k'} \Cov[\phi(s_k^{-}), \phi(s_{k'}^{-})]}_{\Cov_{i \neq j}[s_i, s_j]} + \E[s_i] \E[s_j].
\end{align}
Note that this expression also appears in the variance formula (eq.~\ref{eq:sigprop:mom2_raw}).

We can again assume that pre-activations are (approximately) identically distributed to conclude that
\begin{equation*}
    \forall n \in \{1, 2\} : \forall k : \E\bigl[\phi(s_k^{-})^n\bigr] = \E\bigl[\phi(s_1^{-})^n\bigr].
\end{equation*}
Putting everything together, we arrive at our generalised signal propagation results from equations~\eqref{eq:sigprop:mom1} and~\eqref{eq:sigprop:mom2}:
\begin{align*}
    \E[s_i] &= N \mu_w \E[\phi(s_1^{-})] + \mu_b \\
    \E[s_i s_j] &= \underbrace{\kronecker_{ij} \big(N \sigma_w^2 \E\big[\phi(s_1^{-})^2\big] + \sigma_b^2\big) + \mu_w^2 \sum_{k,k'} \Cov[\phi(s_k^{-}), \phi(s_{k'}^{-})]}_{\Cov[s_i, s_j]} + \E[s_i] \E[s_j].
\end{align*}
Here, we combined equations~\eqref{eq:sigprop:mom2_raw} and \eqref{eq:sigprop:mom_mix_raw} into a single expression using the Kronecker delta, $\kronecker_{ij}$.

The identical distribution of the pre-activations also induces a particular structure in the covariance matrix.
Because the diagonal entries represent the variance, they must have the same value.
The off-diagonal entries happen to be the same --- but typically different from the variance --- as well.
After all, eq.~\eqref{eq:sigprop:mom_mix_raw} is also independent of indices $i$ and $j$.
As a result, the sum over entries in the covariance matrix can be written in terms of these two values:
\begin{align}
    \nonumber
    \sum_{k, k'} \Cov[\phi(s_k^-), \phi(s_{k'}^-)] &= \sum_k \Big(\Var[\phi(s_k^-)] + \sum_{k' \neq k} \Cov[\phi(s_k^-), \phi(s_{k'}^-)]\Big) \\
    \label{eq:covarsum:independent}
    &= N \big(\Var[\phi(s_1^-)] + (N - 1) \Cov[\phi(s_1^-), \phi(s_2^-)]\big).
\end{align}

\subsection{Activation Function Kernels}
\label{app:sigprop:kernels}

Activation functions are not directly affected by the positive weights in \acp{icnn}.
Nevertheless, the propagation through non-linear activation functions also has to be reconsidered for our generalised theory.
The main reason is the difference in covariance structure between the traditional and our generalised signal propagation.


When considering the propagation through non-linearities, the pre-activations are assumed to be Gaussian random variables.
This is typically justified by the central limit theorem, which applies to sums of (weakly) independent random variables.
In the traditional theory (sec.~\ref{app:sigprop:traditional}), where $\mu_w = \mu_b = 0$, it can be verified that pre-activations are uncorrelated:
\begin{align}
    \nonumber
    \E_{i \neq j}[s_i s_j] &= \E\Bigl[\Big(b_i + \sum_k w_{ik} \phi(s_k^{-})\Big) \Big(b_j + \sum_k w_{jk} \phi(s_k^{-})\Big)\Bigr] \\
    \nonumber
    &= \E[b_i b_j] + 2 \E\Bigl[b_i \sum_k w_{jk} \phi(s_k^{-})\Bigr] + \E\Bigl[\sum_{k, k'} w_{ik} w_{jk'} \phi(s_k^{-}) \phi(s_{k'}^{-})\Bigr] \\
    \nonumber
    &= \mu_b^2 + 2 \mu_b \mu_w \sum_k \E[\phi(s_k^{-})] + \mu_w^2 \sum_{k, k'} \E[\phi(s_k^{-}) \phi(s_{k'}^{-})] \\
    \label{eq:standard_sigprop:cov}
    &= 0.
\end{align}
Although this does not imply independence, it is often sufficient to claim weak independence.
Also, empirical investigations typically confirm these assumptions (e.g.~figure~\ref{fig:propagation:nn}).

In contrast, the results of our generalised theory (sec.~\ref{app:sigprop:generalised}) suggest that pre-activations have non-zero correlation.
Therefore, the Gaussian assumption on the pre-activations is hard to justify.
Moreover, this assumption also clearly does not match empirical observations (e.g.~ figure~\ref{fig:propagation:icnn_init}).
Nevertheless, we adopt the Gaussian assumption for the generalised theory due to a lack of better options.

Consider some (non-linear) activation function, $\phi : \reals \to \reals$.
The effects of $\phi$ on signal propagation can be captured by the matrix of mixed moments,
\begin{equation}
    \E_{s_1, s_2 \sim \normaldist(\vec{0}, \mat{\Sigma})}[\phi(s_1) \phi(s_2)],
\end{equation}
where $\mat{\Sigma} = \big(\begin{smallmatrix} 1 & \rho \\ \rho & 1 \end{smallmatrix}\big) \sigma^2$.
Here, $\rho$ is the pair-wise correlation and $\sigma^2$ the variance of pre-activations.
This matrix of mixed moments can also be used to define kernels \citep[see e.g.][]{cho2009kernel}.
Therefore, we will refer to this matrix as the \emph{kernel} of the activation function.
Note that for traditional signal propagation theory, only the diagonal is relevant.


For the sake of example, we show how to compute the kernel for the leaky $\relu$ function \citep{maas13lrelu},
\begin{equation*}
    \leakyrelu : \reals \to \reals : x \mapsto \leakyrelu(x \with \alpha) \begin{cases} x & x \geq 0 \\ \alpha x & x < 0 \end{cases}.
\end{equation*}
We choose the leaky $\relu$ function because it allows for analytical solutions.
Throughout these computations, we assume pre-activations to have zero mean.
Moreover, the covariance between any two features is assumed to be $\mat{\Sigma} = \big(\begin{smallmatrix}1 & \rho \\ \rho & 1\end{smallmatrix}\big) \sigma^2$.
Here $\rho$ is the correlation and $\sigma^2$ is the variance of pre-activations.

Using the kernel for $\relu$ from \citep{daniely16toward, cho2009kernel}, the covariance for leaky $\relu$ is given by
\begin{align}
    \nonumber
    \E[&\leakyrelu(s_1 \with \alpha) \leakyrelu(s_2 \with \alpha)] \\
    \nonumber
    &= \int_{-\infty}^0 \int_{-\infty}^0 \alpha^2 s_1 s_2\, p_\normaldist(s_1, s_2 \with \vec{0}, \mat{\Sigma}) \dx[s_1] \dx[s_2] + \int_{-\infty}^0 \int_0^\infty \alpha s_1 s_2\, p_\normaldist(s_1, s_2 \with \vec{0}, \mat{\Sigma}) \dx[s_1] \dx[s_2] \\
    \nonumber
    &\qquad + \int_0^\infty \int_{-\infty}^0 \alpha s_1 s_2\, p_\normaldist(s_1, s_2 \with \vec{0}, \mat{\Sigma}) \dx[s_1] \dx[s_2] + \int_0^\infty \int_0^\infty s_1\, s_2\, p_\normaldist(s_1, s_2 \with \vec{0}, \mat{\Sigma}) \dx[s_1] \dx[s_2] \\
    \nonumber
    &= (1 + \alpha^2) \E_{\normaldist(\vec{0}, \mat{\Sigma})}[\relu(s_1) \relu(s_2)] - 2 \alpha \E_{\normaldist(\vec{0}, \bar{\mat{\Sigma}})}[\relu(s_1)\relu(s_2)] \\
    &= (1 + \alpha^2) \frac{\sigma^2}{2 \pi} \Big(\sqrt{1 - \rho^2} + \rho \arccos(-\rho) \Big) - 2 \alpha \frac{\sigma^2}{2 \pi} \Big(\sqrt{1 - \rho^2} - \rho \arccos(\rho) \Big) \\
    \nonumber
    &= (1 + \alpha^2) \frac{\sigma^2}{2 \pi} \Big(\sqrt{1 - \rho^2} + \rho \arccos(-\rho) \Big) - 2 \alpha \frac{\sigma^2}{2 \pi} \Big(\sqrt{1 - \rho^2} - \rho \pi + \rho \arccos(-\rho) \Big) \\
    \label{eq:lrelu_kernel}
    &= (1 - \alpha)^2\,\frac{\sigma^2}{2 \pi} \Big(\sqrt{1 - \rho^2} + \rho \arccos(-\rho) \Big) + \alpha \sigma^2 \rho,
\end{align}
where $\bar{\mat{\Sigma}} = \big(\begin{smallmatrix} 1 & -\rho \\ -\rho & 1 \end{smallmatrix}\big) \sigma^2$ appears as a side-effect of flipping the integration boundaries.
Note that this expression also captures the squared mean ($\rho = 0$) and second raw moment ($\rho = 1$).

\subsection{Generalised Backward Propagation}
\label{app:sigprop:backprop}

The generalisation to non-zero mean also applies to backpropagation, where signals are the so-called deltas, the derivatives of the loss, $L$, w.r.t. the pre-activations.
These deltas can be defined using the $\reals^M \to \reals^N$ mapping:
\begin{equation}
    \label{eq:backward:deltas}
    \vec{\delta}^{-} = \der{L}{\vec{s}^{-}} = \phi'\bigl(\vec{s}^{-}\bigr) \mat{W}^\mathsf{T} \vec{\delta}.
\end{equation}
Here, $\mat{W}$ are the weights of the layer that takes $\phi\bigl(\vec{s}^{-}\bigr)$ as inputs.
Note that 

The first two (raw) moments of the deltas from eq.~\eqref{eq:backward:deltas} are:
\begin{align}
    \nonumber
    \E[\delta_j^{-}] &= \E\Bigl[\phi'(s_j^{-}) \sum_k \delta_k w_{kj}\Bigr] \\
    \label{eq:sigprop:backprop:mom1_raw}
    &= \mu_w \E[\phi'(s_j^{-})] \sum_k \E[\delta_k] \\
    \nonumber
    \E\bigl[(\delta_j^{-})^2\bigr] &= \E\biggl[\Big(\phi'(s_j^{-}) \sum_k \delta_k w_{kj}\Big)^2\biggr] \\
    \nonumber
    &= \E\Bigl[\phi'(s_j^{-})^2 \sum_{k,k'} \delta_k \delta_{k'} w_{kj} w_{k'j}\Bigr] \\
    \nonumber
    &= (\sigma_w^2 + \mu_w^2) \E\bigl[\phi'(s_j^{-})^2\bigr] \sum_k \E\bigl[\delta_k^2\bigr] + \mu_w^2 \E\bigl[\phi'(s_j^{-})^2\bigr] \sum_{k,k' \neq k} \E[\delta_k \delta_{k'}] \\
    \label{eq:sigprop:backprop:mom2_raw}
    &= \sigma_w^2 \E\bigl[\phi'(s_j^{-})^2\bigr] \sum_k \E\bigl[\delta_k^2\bigr] + \mu_w^2 \E\bigl[\phi'(s_j^{-})^2\bigr] \sum_{k,k'} \E[\delta_k \delta_{k'}].
\end{align}
The second mixed moment can be derived in a similar way:
\begin{align}
    \nonumber
    \E_{i \neq j}[\delta_i^{-} \delta_j^{-}] &= \E\Bigl[\Big(\phi'(s_i^{-}) \sum_k \delta_k w_{ki}\Big) \Big(\phi'(s_j^{-}) \sum_k \delta_k w_{kj} \Big)\Bigr] \\
    \nonumber
    &= \E\Bigl[\phi'(s_i^{-}) \phi'(s_j^{-}) \sum_{k, k'} w_{ki} w_{k'j} \delta_k \delta_{k'}\Bigr] \\
    \label{eq:sigprop:backprop:mom_mix_raw}
    &= \mu_w^2 \E[\phi'(s_i^{-}) \phi'(s_j^{-})] \sum_{k, k'} \E[\delta_k \delta_{k'}].
\end{align}

From the forward pass (see section~\ref{app:sigprop:generalised}), we know that the moments of pre-activations are independent of their index, i.e. $\forall k: \E\bigl[(s_k^{-})^n\bigr] = \E\bigl[(s_1^{-})^n\bigr]$.
As a result, we also have $\E\bigl[\phi'(s_k^{-})^n\bigr] = \E\bigl[\phi'(s_1^{-})^n\bigr]$ and therefore
\begin{equation*}
    \forall n \in \{1, 2\} : \forall k : \E\bigl[(\delta_k^{-})^n\bigr] = \E\bigl[(\delta_1^{-})^n\bigr].
\end{equation*}
Putting everything together, we obtain the generalised signal propagation for the deltas during backpropagation:
\begin{align*}
    \E[\delta_j^{-}] &= M \mu_w \E[\phi'(s_1^{-})] \E[\delta_1] \\
    \E[\delta_i^{-} \delta_j^{-}] &= \kronecker_{ij} M \sigma_w^2 \E\bigl[\phi'(s_1^{-})^2\bigr] \E\bigl[\delta_1^2\bigr] + \mu_w^2 \E[\phi'(s_i^{-}) \phi'(s_j^{-})] \sum_{k,k'} \E[\delta_k \delta_{k'}],
\end{align*}
where $\kronecker_{ij}$ is the Kronecker delta.

The covariance for the derivative of $\leakyrelu$ under the same assumptions as in section~\ref{app:sigprop:kernels} would be given by
\begin{align}
    \nonumber
    \E[&\leakyrelu'(s_1 \with \alpha) \leakyrelu'(s_2 \with \alpha)] \\
    \nonumber
    &= \int_{-\infty}^0 \int_{-\infty}^0 \alpha^2 p_\normaldist(s_1, s_2 \with \vec{0}, \mat{\Sigma}) \dx[s_1] \dx[s_2] + \int_{-\infty}^0 \int_0^\infty \alpha \, p_\normaldist(s_1, s_2 \with \vec{0}, \mat{\Sigma}) \dx[s_1] \dx[s_2] \\
    \nonumber
    &\qquad + \int_0^\infty \int_{-\infty}^0 \alpha \, p_\normaldist(s_1, s_2 \with \vec{0}, \mat{\Sigma}) \dx[s_1] \dx[s_2] + \int_0^\infty \int_0^\infty p_\normaldist(s_1, s_2 \with \vec{0}, \mat{\Sigma}) \dx[s_1] \dx[s_2] \\
    \nonumber
    &= \big(1 + \alpha^2\big) \frac{1}{2 \pi} \arccos(-\rho) + 2 \alpha \frac{1}{2 \pi} \arccos(\rho) \\
    \nonumber
    &= \big(1 + \alpha^2\big) \frac{1}{2 \pi} \arccos(-\rho) + 2 \alpha \frac{1}{2 \pi} \big(\pi - \arccos(-\rho)\big) \\
    \label{eq:backward:lrelu_kernel}
    &= (1 - \alpha)^2 \frac{1}{2 \pi} \arccos(-\rho) + \alpha
\end{align}

\subsection{Convolutional Layers}
\label{app:sigprop:conv}

Our generalised signal propagation (see section~\ref{app:sigprop:generalised}) is also applicable to convolutional layers.
This section provides the derivations for a one-dimensional cross-correlation,
\begin{equation*}
    s_{ia} = b_i + \sum_{c,k} w_{ick} \phi(s_{c(a + k)}^{-}),
\end{equation*}
but can be generalised in a similar way to higher dimensional operations.
The first two moments (cf.~equations~\ref{eq:sigprop:mom1_raw} and~\ref{eq:sigprop:mom2_raw}) are given by
\begin{align}
    \nonumber
    \E[s_{ia}] &= \E\Bigl[b_i + \sum_{c,k} w_{ick} \phi\bigl(s_{c(a + k)}^{-}\bigr)\Bigr] \\
    \label{eq:sigprop:conv_mom1}
    &= \mu_b + \mu_w \sum_{c,k} \E\bigl[\phi\bigl(s_{c(a + k)}^{-}\bigr)\bigr] \\
    \nonumber
    \E\bigl[s_{ia}^2\bigr] &= \E\biggl[\Big(b_i + \sum_{c,k} w_{ick} \phi\bigl(s_{c(a + k)}^{-}\bigr)\Big)^2\biggr] \\
    \nonumber
    &= \sigma_b^2 + \mu_b^2 + 2 \mu_b \mu_w \sum_{c,k} \E\bigl[\phi\bigl(s_{c(a + k)}^{-}\bigr)\bigr] + \sigma_w^2 \sum_{c,k} \E\Bigl[\phi\bigl(s_{c(a + k)}^{-}\bigr)^2\Bigr] + \mu_w^2 \sum_{c,k} \sum_{c', k'} \E\Bigl[\phi\bigl(s_{c(a + k)}^{-}\bigr) \phi\bigl(s_{c'(a + k')}^{-}\bigr)\Bigr] \\
    \label{eq:sigprop:conv_mom2}
    &= \underbrace{\sigma_b^2 + \sigma_w^2 \sum_{c,k} \E\Bigl[\phi\bigl(s_{c(a + k)}^{-}\bigr)^2\Bigr] + \mu_w^2 \sum_{c,k} \sum_{c', k'} \Cov\Bigl[\phi\bigl(s_{c(a + k)}^{-}\bigr), \phi\bigl(s_{c'(a + k')}^{-}\bigr)\Bigr]}_{\Var[s_{ia}]} + \E[s_{ia}]^2.
\end{align}
For the second mixed moment we find (cf.~eq~\ref{eq:sigprop:mom_mix_raw})
\begin{align}
    \nonumber
    \E_{i \neq j}\bigl[s_{ia} s_{j(a+d)}\bigr] &= \E\Bigl[\Big(b_i + \sum_{c,k} w_{ick} \phi\bigl(s_{c(a + k)}^{-}\bigr)\Big) \Big(b_j + \sum_{c,k} w_{jck} \phi\bigl(s_{c(a + d + k)}^{-}\bigr)\Big)\Bigr] \\
    \nonumber
    &= \mu_b^2 + \mu_b \mu_w \sum_{c,k} \Big(\E\bigl[\phi\bigl(s_{c(a + k)}^{-}\bigr)\bigr] + \E\bigl[\phi\bigl(s_{c(a + d + k)}^{-}\bigr)\bigr]\Big) \\
    \nonumber
    &\qquad + \mu_w^2 \sum_{c,k} \sum_{c',k'} \E\bigl[\phi\bigl(s_{c(a + k)}^{-}\bigr) \phi\bigl(s_{c'(a + d + k')}^{-}\bigr)\bigr] \\
    \label{eq:sigprop:conv_mom_mix1}
    &= \underbrace{\mu_w^2 \sum_{c,k} \sum_{c',k'} \Cov\bigl[\phi\bigl(s_{c(a + k)}^{-}\bigr), \phi\bigl(s_{c'(a + d + k')}^{-}\bigr)\bigr]}_{\Cov[s_{ia}, s_{j(a+d)}]} + \E[s_{ia}] \E[s_{j(a + d)}].
\end{align}
However, this mixed moment does not capture the covariance between elements in the same channel, but on different positions.
Therefore, we additionally include the following result:
\begin{align}
    \nonumber
    \E_{d \neq 0}\bigl[s_{ia} s_{i(a+d)}\bigr] &= \E\Bigl[\Big(b_i + \sum_{c,k} w_{ick} \phi\bigl(s_{c(a + k)}^{-}\bigr)\Big) \Big(b_i + \sum_{c,k} w_{ick} \phi\bigl(s_{c(a + d + k)}^{-}\bigr)\Big)\Bigr] \\
    \nonumber
    &= \sigma_b^2 + \mu_b^2 + \mu_b \mu_w \sum_{c,k} \Big(\E\bigl[\phi\bigl(s_{c(a + k)}^{-}\bigr)\bigr] + \E\bigl[\phi\bigl(s_{c(a + d + k)}^{-}\bigr)\bigr]\Big) \\
    \nonumber
    &\qquad + \sigma_w^2 \sum_{c,k} \E\bigl[\phi\bigl(s_{c(a + k)}^{-}\bigr) \phi\bigl(s_{c(a + d + k)}^{-}\bigr)\bigr] + \mu_w^2 \sum_{c,k} \sum_{c',k'} \E\bigl[\phi\bigl(s_{c(a + k)}^{-}\bigr) \phi\bigl(s_{c'(a + d + k')}^{-}\bigr)\bigr] \\
    \nonumber
    &= \sigma_b^2 + \sigma_w^2 \sum_{c,k} \E\bigl[\phi\bigl(s_{c(a + k)}^{-}\bigr) \phi\bigl(s_{c(a + d + k)}^{-}\bigr)\bigr] \\
    \label{eq:sigprop:conv_mom_mix2}
    &\qquad + \mu_w^2 \sum_{c,k} \sum_{c',k'} \Cov\bigl[\phi\bigl(s_{c(a + k)}^{-}\bigr), \phi\bigl(s_{c'(a + d + k')}^{-}\bigr)\bigr] + \E[s_{ia}] \E[s_{i(a+d)}]
\end{align}

Putting these results together, we can summarise the signal propagation through a convolutional layer as follows:
\begin{align*}
    \E[s_{ia}] &= \mu_b + \mu_w \sum_{c,k} \E\bigl[\phi\bigl(s_{c(a + k)}^{-}\bigr)\bigr] \\
    \E[s_{ia} s_{j(a + d)}] &= \kronecker_{ij} \bigg(\sigma_b^2 + \sigma_w^2 \sum_{c,k} \E\bigl[\phi\bigl(s_{c(a + k)}^{-}\bigr) \phi\bigl(s_{c(a + d + k)}^{-}\bigr)\bigr]\bigg) \\
    &\qquad + \mu_w^2 \sum_{c,k} \sum_{c',k'} \Cov\bigl[\phi\bigl(s_{c(a + k)}^{-}\bigr), \phi\bigl(s_{c'(a + d + k')}^{-}\bigr)\bigr] + \E[s_{ia}] \E[s_{j(a+d)}]
\end{align*}
Here, we use the Kronecker delta to combine equations~\eqref{eq:sigprop:conv_mom2},~\eqref{eq:sigprop:conv_mom_mix1} and~\eqref{eq:sigprop:conv_mom_mix2} into one equation.

\section{Deriving the Initialisation}
\label{app:icnn_init}

This section aims to show how the \ac{icnn} initialisation can be derived from our generalised signal propagation theory.

\subsection{Zero Mean Pre-Activations}
\label{app:icnn_init:preact_mean}

The goal of a good initialisation is to have pre-activations with similar statistics in every layer.
In the traditional theory, mean and covariance are zero in every layer by default.
Therefore, variance is the only statistic that needs to be controlled.
This is why traditional initialisation methods \citep{lecun98efficient, glorot10understanding, he15delving} typically focus on the standard deviation of weights.

Consider the equation for pre-activations from eq.~\eqref{eq:forward:pre-acts}.
Assuming $\phi = \leakyrelu,$ we can use the kernel from eq.~\eqref{eq:lrelu_kernel}.
Plugging the result for $\rho = 1$ into eq.~\eqref{eq:standard_sigprop:mom2} we obtain the traditional variance propagation
\begin{equation*}
    \E\bigl[s_i^2 \bigr] = N \sigma_w^2 (1 + \alpha^2) \frac{1}{2} \E\bigl[(s_1^{-})^2\bigr].
\end{equation*}
By requiring that $\E\bigl[s_i^2 \bigr] = \E\bigl[(s_1^{-})^2\bigr] = \sigma_*^2 > 0$,
we obtain the fixed point equation
\begin{equation*}
    \sigma_*^2 = N \sigma_w^2 (1 + \alpha^2) \frac{1}{2} \sigma_*^2,
\end{equation*}
which can be solved for $\sigma_w^2$ to obtain the initialisation variance for (regular) $\leakyrelu$ networks:
\begin{equation*}
    \sigma_w^2 = \frac{2}{1 + \alpha^2} \frac{1}{N}.
\end{equation*}
Here, $2 / (1 + \alpha^2)$ is sometimes also referred to as the \emph{gain} of the initialisation \citep[cf.][]{saxe14exact}.

We adopt a similar approach for our generalised propagation results.
In contrast to the traditional setting, however, we can not directly rely on the activation kernels.
After all, the pre-activations in our general setting are not automatically zero.

\paragraph{Derivation of Bias Mean} 
By setting eq.~\eqref{eq:sigprop:mom1} to zero and solving for $\mu_b$, we get
\begin{equation*}
    \mu_b = -N \mu_w \E[\phi(s_1^{-})].
\end{equation*}
This also makes it easier to use the activation kernels, which assume zero mean inputs.
Note, however, that the activation kernels theoretically do not apply to our setting (see sec.~\ref{app:sigprop:kernels}).
Nevertheless, we assume that the activation kernels approximate the actual dynamics well enough.
Plugging in the square root of the $\leakyrelu$ kernel (eq.~\ref{eq:lrelu_kernel}) with $\rho = 0$, we obtain the bias mean for $\phi = \leakyrelu$:
\begin{align}
    \nonumber
    \mu_b &= -N \mu_w (1 - \alpha) \sqrt{\frac{1}{2 \pi} \Big(\E\bigl[(s_1^{-})^2\bigr] - \E[s_1^{-}]^2\Big)} \\
    \label{eq:init:bias_mean}
    &= -N \mu_w (1 - \alpha) \sqrt{\frac{1}{2 \pi} \E\bigl[(s_1^{-})^2\bigr]}.
\end{align}

\subsection{Variance and Covariance Fixed Points}
\label{app:icnn_init:fixpoints}

To obtain fixed points for the variance and covariance, we start from equations~\eqref{eq:sigprop:mom2_raw} and~\eqref{eq:sigprop:mom_mix_raw}, respectively.
Assuming identically distributed pre-activations, we can use equations~\eqref{eq:varsum:independent} and~\eqref{eq:covarsum:independent}, to get expressions without sums:
\begin{align*}
    \E\bigl[s_i^2\bigr] &= \sigma_b^2 + \sigma_w^2 N \E\bigl[\phi(s_1^{-})^2\bigr] + \E_{i \neq j}[s_i s_j] \\
    \E_{i \neq j}[s_i s_j] &= \mu_w^2 N \Big( \Var\bigl[\phi(s_1^{-})^2\bigr] + (N - 1) \Cov\bigl[\phi(s_1^{-}), \phi(s_2^{-})\bigr] \Big)
\end{align*}
For $\phi = \leakyrelu$, we can use the kernel from eq.~\eqref{eq:lrelu_kernel} with $\rho = 1$ to work out the moments for the variance,
\begin{equation}
    \label{eq:init:var_prop}
    \E\bigl[s_i^2\bigr] = \sigma_b^2 + \sigma_w^2 N (1 + \alpha^2) \frac{1}{2} \E\bigl[(s_1^{-})^2\bigr] + \E_{i \neq j}[s_i s_j],
\end{equation}
and with $\rho = c^{-} = \Corr[s_1^{-}, s_2^{-}] = \E[s_1^{-} s_2^{-}] / \E\bigl[(s_1^{-})^2\bigr]$ to obtain the covariance
\begin{align*}
    \nonumber
    \E_{i \neq j}[s_i s_j] &= \mu_w^2 N \Big( \E\bigl[\leakyrelu(s_1^{-})^2\bigr] + (N - 1) \E\bigl[\leakyrelu(s_1^{-})  \leakyrelu(s_2^{-})\bigr] - N \E\bigl[\leakyrelu(s_1^{-})\bigr]^2 \Big) \\
    \nonumber
    &= \mu_w^2 N \Bigg( \frac{1 + \alpha^2}{2} \E\bigl[(s_1^{-})^2\bigr] - N \frac{(1 - \alpha)^2}{2 \pi} \E\bigl[(s_1^{-})^2\bigr] \\
    \nonumber
    &\qquad + (N - 1) \bigg( \frac{(1 - \alpha)^2}{2 \pi} \E\bigl[(s_1^{-})^2\bigr] \Big(\sqrt{1 - (c^{-})^2} + c^{-} \arccos\bigl(-c^{-}\bigr)\Big) + \alpha \E\bigl[(s_1^{-})^2\bigr] c^{-} \bigg) \Bigg) \\
    \nonumber
    &= \mu_w^2 \frac{N}{2 \pi} \E\bigl[(s_1^{-})^2\bigr] \Bigg( (1 + \alpha^2) \pi - N (1 - \alpha)^2 \\
    &\qquad + (N - 1) \bigg((1 - \alpha)^2 \Big(\sqrt{1 - (c^{-})^2} + c^{-} \arccos\bigl(-c^{-}\bigr)\Big) + 2 \pi \alpha c^{-}\bigg) \Bigg)
\end{align*}
For the sake of readability, we introduce $f_\mathrm{c} : [-1, 1] \to \reals$ to denote
\begin{equation}
    \label{eq:init:c_func}
    f_\mathrm{c}(\rho) = \frac{N}{2 \pi} \Bigg( \big(1 + \alpha^2\big) \pi - N (1 - \alpha)^2 + (N - 1) \bigg((1 - \alpha)^2 \Big(\sqrt{1 - \rho^2} + \rho \arccos\bigl(-\rho\bigr)\Big) + 2 \pi \alpha \rho\bigg) \Bigg),
\end{equation}
such that the covariance can be written more compactly as
\begin{equation}
    \label{eq:init:covariance_compact}
    \E_{i \neq j}[s_i s_j] = \mu_w^2 \E\bigl[(s_1^{-})^2\bigr] f_\mathrm{c}\bigl(\Corr[s_1^{-}, s_2^{-}]\bigr)
\end{equation}

\paragraph{Derivation of Bias and Weight Variances}
The fixed point equation for the variance (eq.~\ref{eq:init:var_prop}) with $\E\bigl[s_i^2\bigr] = \E\bigl[(s_1^{-})^2\bigr] = \sigma_*^2$ can be written as
\begin{equation*}
    \sigma_*^2 = \sigma_b^2 + \sigma_w^2 N (1 + \alpha^2) \frac{1}{2} \sigma_*^2 + \sigma_*^2 \mu_w^2 f_\mathrm{c}\bigl(\Corr[s_1^{-}, s_2^{-}]\bigr),
\end{equation*}
where $f_\mathrm{c}$ is taken from equation~\eqref{eq:init:c_func}.
Note that we retrieve eq.~\eqref{eq:icnn:var} in the main text from this result by setting $\alpha = 0$ and replacing $\mu_w^2 f_\mathrm{c}\bigl(\Corr[s_1^{-}, s_2^{-}]\bigr)$ by the correlation fixed point, $\rho_*$.
We elaborate on the latter when discussing the fixed point equation for the correlation.
Solving the fixed point equation for $\sigma_w^2$, we find
\begin{equation}
    \label{eq:init:weight_var_bias}
    \sigma_w^2 = \frac{2}{1 + \alpha^2} \frac{1}{N}\Big(1 - \frac{\sigma_b^2}{\sigma_*^2} - \mu_w^2 f_\mathrm{c}\bigl(\Corr[s_1^{-}, s_2^{-}]\bigr)\Big).
\end{equation}
By initialising the bias vector with a constant value, such that $\sigma_b^2 = 0$, the expression above becomes independent of the fixed point $\sigma_*^2$:
\begin{equation}
    \label{eq:init:weight_var_clean}
    \sigma_w^2 = \frac{2}{1 + \alpha^2} \frac{1}{N}\Big(1 - \mu_w^2 f_\mathrm{c}\bigl(\Corr[s_1^{-}, s_2^{-}]\bigr)\Big).
\end{equation}
Using similar substitutions as before, we obtain the result from the main text (eq.~\ref{eq:weightdist:var}).
A further simplification is possible by revisiting eq.~\eqref{eq:init:covariance_compact} and observing that if $\E\bigl[s_i^2\bigr] = \E\bigl[(s_1^{-})^2\bigr]$, we have $\Corr[s_i, s_j] = \mu_w^2 f_\mathrm{c}\bigl(\Corr[s_1^{-}, s_2^{-}]\bigr)$, such that
\begin{equation}
    \label{eq:init:weight_var_clear}
    \sigma_w^2 = \frac{2}{1 + \alpha^2} \frac{1}{N}\Big(1 - \Corr[s_i, s_j]\Big).
\end{equation}

\paragraph{Derivation of Weight Mean}
Because eq.~\eqref{eq:init:covariance_compact} is mainly a function of correlation, we consider a fixed point equation in terms of correlation rather than covariance.
Concretely, we set $\Corr[s_i, s_j] = \Corr[s_1^{-}, s_2^{-}] = \rho_*$ to obtain the fixed point equation
\begin{equation*}
    \rho_* = \mu_w^2 \frac{\E\bigl[(s_1^{-})^2\bigr]}{\E\bigl[s_1^2\bigr]} f_\mathrm{c}(\rho_*),
\end{equation*}
where we used the identical distribution assumption as follows
\begin{equation*}
    \Corr[s_i, s_j] = \frac{\E_{i \neq j}\bigl[s_i s_j\bigr]}{\sqrt{\E\bigl[s_i^2\bigr] \E\bigl[s_j^2\bigr]}} = \frac{\E_{i \neq j}\bigl[s_i s_j\bigr]}{\E\bigl[s_1^2\bigr]}.
\end{equation*}
Note that for $\alpha = 0$ and $\E\bigl[s_1^2\bigr] = \E\bigl[(s_1^{-})^2\bigr]$, this fixed point equation corresponds to eq.~\eqref{eq:icnn:corr} in the main text.
Solving the fixed point equation for $\mu_w^2$, we find
\begin{equation}
    \label{eq:init:weight_mean_clear}
    \mu_w^2 = \frac{\E\bigl[s_1^2\bigr]}{\E\bigl[(s_1^{-})^2\bigr]} \rho_* \, f_\mathrm{c}(\rho_*)^{-1}.
\end{equation}
Under similar conditions as previously stated, this gives us the result from the main text (eq.~\ref{eq:weightdist:mean}).
We can additionally plug in eq.~\eqref{eq:init:var_prop} into this result to get rid of $\E\bigl[s_1^2\bigr]$:
\begin{equation*}
    \mu_w^2 = \bigg(\frac{\sigma_b^2}{\E\bigl[(s_1^{-})^2\bigr]} + \sigma_w^2 N \,  \frac{1 + \alpha^2}{2} + \rho_*\bigg) \rho_* \, f_\mathrm{c}(\rho_*)^{-1}.
\end{equation*}
Assuming $\sigma_b^2 = 0$ again, we obtain an expression that is independent of $\E\bigl[(s_1^{-})^2\bigr]$:
\begin{equation}
    \label{eq:init:weight_mean_clean}
    \mu_w^2 = \Big(\sigma_w^2 N \, \frac{1 + \alpha^2}{2} + \rho_*\Big) \rho_* \, f_\mathrm{c}(\rho_*)^{-1}.
\end{equation}

\subsection{Initialisation Parameters}
\label{app:icnn_init:params}

Putting everything together from sections~\ref{app:icnn_init:preact_mean} and~\ref{app:icnn_init:fixpoints}, we obtain an initialisation for fully-connected layers with features produced by a $\leakyrelu$ non-linearity.
The initialisation parameters are (see equations~\ref{eq:init:bias_mean},~\ref{eq:init:weight_var_clean} and~\ref{eq:init:weight_mean_clean}):
\begin{align*}
    \mu_w &= \pm \sqrt{\bigg(\sigma_w^2 N \, \frac{1 + \alpha^2}{2} + \rho_*\bigg) \rho_* \, f_\mathrm{c}(\rho_*)^{-1}} &
    \sigma_w^2 &= \frac{2}{1 + \alpha^2} \frac{1}{N} \Big(1 - \mu_w^2 f_\mathrm{c}\bigl(\Corr[s_1^{-}, s_2^{-}]\bigr)\Big) \\
    \mu_b &= -N \mu_w (1 - \alpha) \sqrt{\frac{1}{2 \pi} \E\bigl[(s_1^{-})^2\bigr]} &
    \sigma_b^2 &= 0.
\end{align*}

Note that the mean for the weights is the only result that directly depends on the fixed point.
However, because these results are strongly connected, they have to be considered simultaneously.
Practically, this means that we have to focus on the joint fixed point $(\sigma_*, \rho_*)$ instead of the individual parts.
Considering equations~\eqref{eq:init:weight_var_clear} and~\eqref{eq:init:weight_mean_clear},
it is clear that the joint fixed point can be obtained with the following formulations
\begin{align}
    \label{eq:init:weight_pars}
    \mu_w &= \pm \sqrt{\rho_* \, f_\mathrm{c}(\rho_*)^{-1}} &
    \sigma_w^2 &= \frac{2}{1 + \alpha^2} \frac{1}{N} (1 - \rho_*) \\
    \label{eq:init:bias_pars}
    \mu_b &= -N \mu_w (1 - \alpha) \sqrt{\frac{\sigma_*^2}{2 \pi}} &
    \sigma_b^2 &= 0.
\end{align}

Our initialisation parameters still depend on the choice of the joint fixed point $\sigma_*^2$ and $\rho_*$.
Normally, $\sigma_*^2 = 1$ is assumed because that is how the input data is typically normalised.
In general, we can assume $\sigma_*^2$ to be the variance of the input data.
The ideal scenario for correlation is to have $\rho_* = 0$, i.e.~uncorrelated features.
However, when plugging this into our initialisation parameters, we retrieve the traditional setting from section~\ref{app:sigprop:traditional}, where $\mu_w = 0$.
In other words, uncorrelated features are only possible if $\mu_w = 0$.
Also, $\abs{\rho_*} < 1$ is desirable, because $\rho = 1$ practically corresponds to all values being the same.
Therefore, we (arbitrarily) choose to set $\rho_* = \frac{1}{2}$.
One argument in favour of this choice is that eq.~\eqref{eq:init:c_func} has a relatively nice solution
\begin{align*}
    f_\mathrm{c}\biggl(\frac{1}{2}\biggr) &= \frac{N}{2 \pi} \Bigg( \big(1 + \alpha^2\big) \pi - N (1 - \alpha)^2 + (N - 1) \bigg((1 - \alpha)^2 \Big(\frac{\sqrt{3}}{2} + \frac{\pi}{3}\Big) + \pi \alpha\bigg) \Bigg) \\
    &= \frac{N}{12 \pi} \bigg( 6 \big(1 + \alpha^2\big) \pi - 6 N (1 - \alpha)^2 + (N - 1) \Big((1 - \alpha)^2 \big(3 \sqrt{3} + 2 \pi\big) + 6 \pi \alpha\Big) \bigg) \\
    &= \frac{N}{12 \pi} \bigg( 6 (1 - \alpha)^2 \pi + 12 \alpha \pi - 6 N (1 - \alpha)^2 + (N - 1) (1 - \alpha)^2 \big(3 \sqrt{3} + 2 \pi\big) + 6 (N - 1) \pi \alpha \bigg) \\
    &= \frac{N}{12 \pi} \bigg( (1 - \alpha)^2 \Big(6\pi - 6N + (N - 1) \big(3 \sqrt{3} + 2 \pi\big)\Big) + 6 (N + 1) \pi \alpha \bigg) \\
    &= \frac{N}{12 \pi} \bigg( (1 - \alpha)^2 \Big(6 (\pi - 1) + (N - 1) \big(3 \sqrt{3} + 2 \pi - 6\big)\Big) + 6 (N + 1) \pi \alpha \bigg).
\end{align*}

Filling out the fixed point $(\sigma_*^2, \rho_*) = (1, \frac{1}{2})$, we obtain the final initialisation parameters for \acp{icnn} with $\leakyrelu$ activation functions:
\begin{align*}
    \mu_w &= \pm \sqrt{\frac{6 \pi}{N \bigg( (1 - \alpha)^2 \Big(6 \pi - 6 N + (N - 1) \big(3 \sqrt{3} + 2 \pi\big)\Big) + 6 (N + 1) \pi \alpha \bigg)}} &
    \sigma_w^2 &= \frac{1}{1 + \alpha^2} \frac{1}{N} \\
    \mu_b &= \mp \sqrt{\frac{3 N (1 - \alpha)^2}{(1 - \alpha)^2 \Big(6 \pi - 6 N + (N - 1) \big(3 \sqrt{3} + 2 \pi\big)\Big) + 6 (N + 1) \pi \alpha}} &
    \sigma_b^2 &= 0.
\end{align*}
Note again that these results correspond to the results in the main paper where $\alpha = 0$.

\subsection{Fixed Point Stability}
\label{app:icnn_init:fixpoint_stability}

The goal of this section is to establish the stability of the joint fixed point we used for our initialisation from section~\ref{app:icnn_init:params}.
Therefore, we analyse the Jacobian of the mapping
\begin{equation*}
    f_* : \reals^+ \times [-1, 1] \to \reals^+ \times [-1, 1] : (\sigma^2, \rho) \mapsto \big(\sigma_b^2 + \sigma_w^2 N \big(1 + \alpha^2\big) \frac{1}{2} \sigma^2 + \sigma^2 \mu_w^2 f_\mathrm{c}(\rho), \mu_w^2 f_\mathrm{c}(\rho)\big),
\end{equation*}
which combines equations~\eqref{eq:init:var_prop} and~\eqref{eq:init:covariance_compact} with $f_\mathrm{c}$ from eq.~\eqref{eq:init:c_func}.
Filling out our initialisation parameters from equations~\eqref{eq:init:weight_pars} and~\eqref{eq:init:bias_pars}, we obtain
\begin{equation}
    f_*(\sigma^2, \rho) = \bigg(1 - \rho_* + \sigma^2 \rho_* \frac{f_\mathrm{c}(\rho)}{f_\mathrm{c}(\rho_*)}, \rho_* \frac{f_\mathrm{c}(\rho)}{f_\mathrm{c}(\rho_*)}\bigg).
\end{equation}

The stability of $f_*$ can be analysed by computing its Jacobian
\begin{equation*}
    \mat{\mathcal{J}}_{f_*}(\sigma^2, \rho) = \begin{bmatrix}
        \rho_* \frac{f_\mathrm{c}(\rho)}{f_\mathrm{c}(\rho_*)} & \sigma^2 \rho_* \frac{f_\mathrm{c}'(\rho)}{f_\mathrm{c}(\rho_*)} \\
        0 & \rho_* \frac{f_\mathrm{c}'(\rho)}{f_\mathrm{c}(\rho_*)}
    \end{bmatrix},
\end{equation*}
where
\begin{align*}
    f_\mathrm{c}'(\rho) &= \frac{N}{2 \pi} (N - 1) (1 - \alpha)^2 \der{}{\rho}\Bigl(\sqrt{1 - \rho^2} + \rho \arccos(-\rho)\Bigr) + \frac{N}{2 \pi} (N - 1) 2 \pi \alpha \\
    &= \frac{N}{2 \pi} (N - 1) (1 - \alpha)^2 \Bigg(\frac{-\rho}{\sqrt{1 - \rho^2}} + \arccos(-\rho) + \rho \frac{1}{\sqrt{1 - \rho^2}}\Bigg) + N (N - 1) \alpha \\
    &= \frac{N}{2 \pi} (N - 1) (1 - \alpha)^2 \arccos(-\rho) + N (N - 1) \alpha.
\end{align*}
The eigenvalues of the Jacobian are the roots of the characteristic polynomial
\begin{equation*}
    \abs{\lambda \mat{I} - \mat{\mathcal{J}}_{f_*}(\sigma^2, \rho)} = \bigg(\lambda - \rho_* \, \frac{f_\mathrm{c}(\rho)}{f_\mathrm{c}(\rho_*)}\bigg) \bigg(\lambda - \rho_* \, \frac{f_\mathrm{c}'(\rho)}{f_\mathrm{c}(\rho_*)}\bigg),
\end{equation*}
which happen to be the entries on the diagonal:
\begin{align*}
    \lambda_1 &= \rho_* \, \frac{f_\mathrm{c}(\rho)}{f_\mathrm{c}(\rho_*)} &
    \lambda_2 &= \rho_* \, \frac{f_\mathrm{c}'(\rho)}{f_\mathrm{c}(\rho_*)}.
\end{align*}

Evaluating the eigenvalues of the Jacobian at the joint fixed point, $(\sigma^2_*, \rho_*)$, we find
\begin{align*}
    \lambda_1 &= \rho_* &
    \lambda_2 &= \rho_* \, \frac{f_\mathrm{c}'(\rho_*)}{f_\mathrm{c}(\rho_*)}.
\end{align*}
Setting $\rho_* = \frac{1}{2}$ and $\alpha = 0$, this becomes
\begin{equation*}
    \lambda_2 = \frac{(N - 1) 2 \pi}{6 \pi - 6 + (N - 1) (3 \sqrt{3} + 2 \pi - 6)},
\end{equation*}
which is only less than one if
\begin{equation*}
    N < 1 + \frac{2 \pi - 2}{2 - \sqrt{3}} \approx 17.
\end{equation*}
As a result, the fixed point $(\sigma^2_*, \rho_*) = (1, \frac{1}{2})$ is not stable in practical settings.

\section{Additional Experiments}
\label{app:experiments}

This section presents experimental details and results that did not make it into the main paper.

\subsection{Computing resources and budget}

For our experiments, we had access to server infrastructure with multiple {GPU}s.
We mainly used {NVIDIA} Titan~V and {NVIDIA} {RTX}~{2080Ti} graphics cards with approximately 12~{GB} of memory for the computer vision experiments.
Experiments were run in parallel on up to 8 {GPU}s.
To maximise {GPU} usage, repetitions also ran in parallel on single cards.
The {Tox21} experiments were run on a desktop {PC} with one {NVIDIA} {GTX}~{1070Ti}, which has 8~{GB} of on-board memory.

A single run for the experiments in Fig.~\ref{fig:learn_curves}, Fig.~\ref{fig:learn_curves_depth}, and Fig.~\ref{fig:learn_curves_exp} took approximately one minute and used no more than 2~{GB} of {GPU} memory on {MNIST}.
For {CIFAR10} and {CIFAR100}, a single run took up to four minutes and used up to 4~{GB} of {GPU} memory.
As a result, an (over-)estimate for the compute time for Figure~\ref{fig:learn_curves} is 6~hours or 18~hours for Figure~\ref{fig:learn_curves_depth}.
All the repetitions for the {Tox21} results in Table~\ref{tab:tox21results}, were obtained in 2~hours --- i.e.~30~minutes per model --- using a little over 500~{MB} of {GPU} memory.

The hyper-parameter search that lead to Table~\ref{tab:hparams}, required the most compute.
For these estimates, we processed the timestamps for each run in our logs.
Since we did not log memory consumption, we can not reliably report this data.
The {MNIST} non-convex baseline model search required approximately 24~hours with a median run-time of less than four minutes.
The search for the non-convex model on {CIFAR10} required 9~hours with a median run-time of approximately two minutes.
Note that five runs ran simultaneously (on the same card) for {MNIST}, but only three runs ran at the same time for {CIFAR10}.
The search for ``\acs{icnn}'' took 11~hours for {MNIST} and 44~hours for {CIFAR10}.
The search for ``\acs{icnn}~+~skip'' took 14~hours on both {MNIST} and {CIFAR10}.
The search for ``\acs{icnn}~+~init'' took 45~hours on {MNIST} and 40~hours for {CIFAR10}.
This gives a total of 201~hours wall-clock computation time for the hyper-parameter search.
Training the resulting ten repetitions for the eight resulting models required an additional 2.5~hours for {MNIST} and 3.5~hours for {CIFAR10} to obtain figure~\ref{fig:fine_tuned}.

\subsection{Computer Vision Benchmarking Datasets}
\label{app:experiments:init}

\begin{figure}
    \centering
    \includegraphics[width=\linewidth]{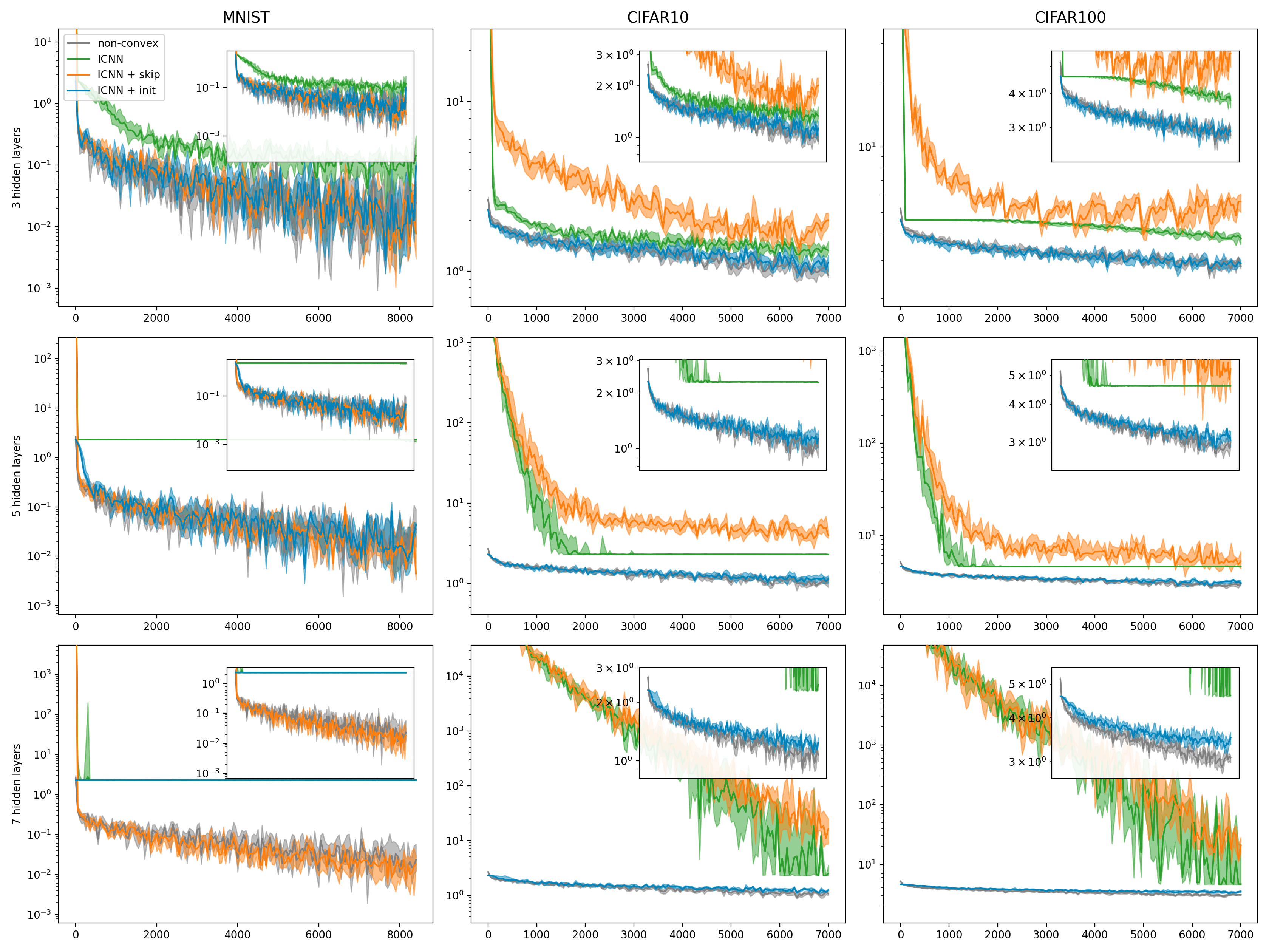}
    \caption{Training loss curves of \ac{icnn} variants with the same architecture with three, five and seven hidden layers on the {MNIST}, {CIFAR10} and {CIFAR100} datasets. ``\acs{icnn}'' is an \acl{icnn} with He initialisation. ``\acs{icnn}~+~skip'': same settings but with skip-connections. ``\acs{icnn}~+~init'': our principled initialisation for \acp{icnn} w/o skip-connections. ``non-convex'': a traditional non-convex network. The median performance over ten runs is displayed, shaded regions represent the inter-quartile range. The inset figures provide a view of the loss curves zoomed in. }
    \label{fig:learn_curves_depth}
\end{figure}

In this section, we provide additional details on the experiments on computer vision datasets.
Figure~\ref{fig:learn_curves_depth} extends Figure~\ref{fig:learn_curves} from the main paper with learning curves for networks with three and seven hidden layers.
Table~\ref{tab:hparams} lists the best hyper-parameters for the search behind Figure~\ref{fig:fine_tuned}.
Table~\ref{tab:hparam_space} lists the options for the different hyper-parameters that we used.

\begin{table}
    \centering
    \caption{Hyper-parameter search space for results in Table~\ref{tab:hparams}}
    \label{tab:hparam_space}
    \begin{tabular}{lcl}
        \toprule
        pre-processing &  & \{none, {PCA}, {ZCA}\} \\
        learning rate &  & $\{10^{-2}, 10^{-3}, 10^{-4}\}$ \\
        $L_2$ regularisation &  & $\{0, 10^{-2}\}$ \\
         & 1 hidden & \{(1k), (10k)\} \\
        layer-widths & 2 hidden & \{(1k, 1k), (1k, 10k), (10k, 1k), (10k, 10k)\} \\
         & 3 hidden & \{(1k, 1k, 1k), (1k, 10k, 1k), (10k, 1k, 10k), (10k, 10k, 10k)\} \\
         \bottomrule
    \end{tabular}
\end{table}

\begin{table}
    \centering
    \caption{
        Optimal hyper-parameters for each configuration in Figure~\ref{fig:fine_tuned}. 
        The epoch column reports the epoch in which the validation accuracy was highest.
    }
    \label{tab:hparams}
    \begin{tabular}{lllccccr}
        \toprule
        &           & architecture                & $\eta$ & $L_2$ & pre-processing & epoch & accuracy \\ \midrule
        \multirow{4}{1em}{\rotatebox{90}{MNIST}} 
        & non-convex  & (784, 1\,000, 1\,000, 10)          & 1e-4   & 0.01  & {ZCA}   & 25 & 98.62\%  \\
        & \ac{icnn} & (784, 10\,000, 10)                   & 1e-4   & 0.01  & {PCA}   & 24 & 98.28\%  \\
        & \ac{icnn} + skip & (784, 10\,000, 10)            & 1e-3   & 0.00  & none    & 24 & 98.30\% \\
        & \ac{icnn} + init & (784, 10\,000, 10)            & 1e-4   & 0.00  & {ZCA}   & 15 & 98.27\%  \\
        \midrule
        \multirow{4}{1em}{\rotatebox{90}{CIFAR10}}
        & non-convex  & (3072, 10\,000, 1\,000, 10)  & 1e-4   & 0.01  & {ZCA}    & 18 & 55.92\%  \\
        & \ac{icnn} & (3072, 1\,000, 10)             & 1e-4   & 0.00  & none     & 19 & 54.74\%  \\
        & \ac{icnn} + skip & (3072, 1\,000, 10)      & 1e-4   & 0.01  & none     & 19 & 52.47\%  \\
        & \ac{icnn} + init & (3072, 1\,000, 10)      & 1e-4   & 0.00  & none     & 17 & 55.64\%  \\
        \bottomrule
    \end{tabular}
\end{table}

\subsection{Random Bias Initialisation}
\label{app:experiments:rand_bias}

In section~\ref{sec:icnn_weightdist}, we chose to set $\sigma_b^2 = 0$ for simplicity.
However, it is also possible to derive an initialisation with $\sigma_b^2 > 0$.
Practically, this means that we can initialise the biases with random samples from a Gaussian distribution with mean $\mu_b$  as in eq.~\eqref{eq:bias:mean} and $\sigma_b^2 > 0$.
However, due to eq.~\eqref{eq:icnn:var} we know that this will also affect the initialisation of the synaptic weight parameters.

For the variance of the weight variance to be positive we must consider the following inequality (from eq.~\ref{eq:init:weight_var_bias}):
\begin{align*}
    \sigma_w^2 = \frac{2}{1 + \alpha^2} \frac{1}{N} \Big(1 - \frac{\sigma_b^2}{\sigma_*^2} - \mu_w^2 f_\mathrm{c}\bigl(\Corr[s_1^{-}, s_2^{-}]\bigr)\Big) & > 0 \\
    \Leftrightarrow 1 - \mu_w^2 f_\mathrm{c}\bigl(\Corr[s_1^{-}, s_2^{-}]\bigr) &> \frac{\sigma_b^2}{\sigma_*^2} \\
    \Leftrightarrow \sigma_*^2 \Big(1 - \mu_w^2 f_\mathrm{c}\bigl(\Corr[s_1^{-}, s_2^{-}]\bigr)\Big) &> \sigma_b^2.
\end{align*}
Assuming we are working in a fixed point regime, we obtain the following simplified constraint: $0 \leq \sigma_b^2 < \sigma_*^2 (1 - \rho_*)$.
Based on this constraint, we propose to specify the bias variance using $\sigma_b^2 = \beta (1 - \rho_*) \sigma_*^2, \beta \in \left[0, 1\right)$.
As a result, we obtain the following expression for the variance of the weights:
\begin{equation*}
    \sigma_w^2 = \frac{2}{1 + \alpha^2} \frac{1}{N} \Big(1 - \beta (1 - \rho^*) - \mu_w^2 f_c\bigl(\Corr[s_1^{-}, s_2^{-}]\bigr)\Big).
\end{equation*}
Combining this result with equations~\eqref{eq:init:bias_mean} and~\eqref{eq:init:weight_mean_clear} under similar conditions as in section~\ref{app:icnn_init:params}, we obtain the following initialisation parameters:
\begin{align}
    \label{eq:init:weight_pars_with_bias}
    \mu_w &= \pm \sqrt{\rho_* \, f_\mathrm{c}(\rho_*)^{-1}} &
    \sigma_w^2 &= \frac{2}{1 + \alpha^2} \frac{1}{N} (1 - \rho_*) (1 - \beta) \\
    \label{eq:init:bias_pars_with_bias}
    \mu_b &= -N \mu_w (1 - \alpha) \sqrt{\frac{\sigma_*^2}{2 \pi}} &
    \sigma_b^2 &= \beta (1 - \rho_*) \sigma_*^2.
\end{align}
Note that this introduces an additional hyper-parameter, $\beta \in [0, 1)$.

\begin{figure}
    \centering%
    \raisebox{17ex}{
        \begin{tabular}{l}
            \textsf{pre-activation} \\
            \textsf{distributions} \\[1.7em]
            \textsf{weight signs} \\
            {\color[HTML]{9ebeff} $\mathsf{{} < 0}$} or {\color[HTML]{d65244} $\mathsf{{} \geq 0}$} \\[1.7em]
            \textsf{feature} \\
            \textsf{correlations} \\[1.5em]
        \end{tabular}%
    }
    \hfill
    \begin{subfigure}{.35\textwidth}
        \centering
        \includegraphics[width=\linewidth]{figures/propagation_diff_init.pdf}
        \caption{\acs{icnn} + init}
        \label{fig:propagation_bias:icnn_init}
    \end{subfigure}
    \hfill
    \begin{subfigure}{.35\textwidth}
        \centering
        \includegraphics[width=\linewidth]{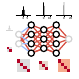}
        \caption{\acs{icnn} + bias-init}
        \label{fig:propagation_bias:icnn_rbias}
    \end{subfigure}
    \caption{
        Illustration of the effects due to good signal propagation in hidden layers.
        Blue and red connections depict negative and positive weights, respectively.
        The distributions of pre-activations in each layer are shown on the top.
        On the bottom, the feature correlation matrices in each layer are displayed.
        The input distribution is depicted by the small elements on the left.
    }
    \label{fig:propagation_bias}
\end{figure}

\begin{figure}
    \centering
    \includegraphics[width=\linewidth]{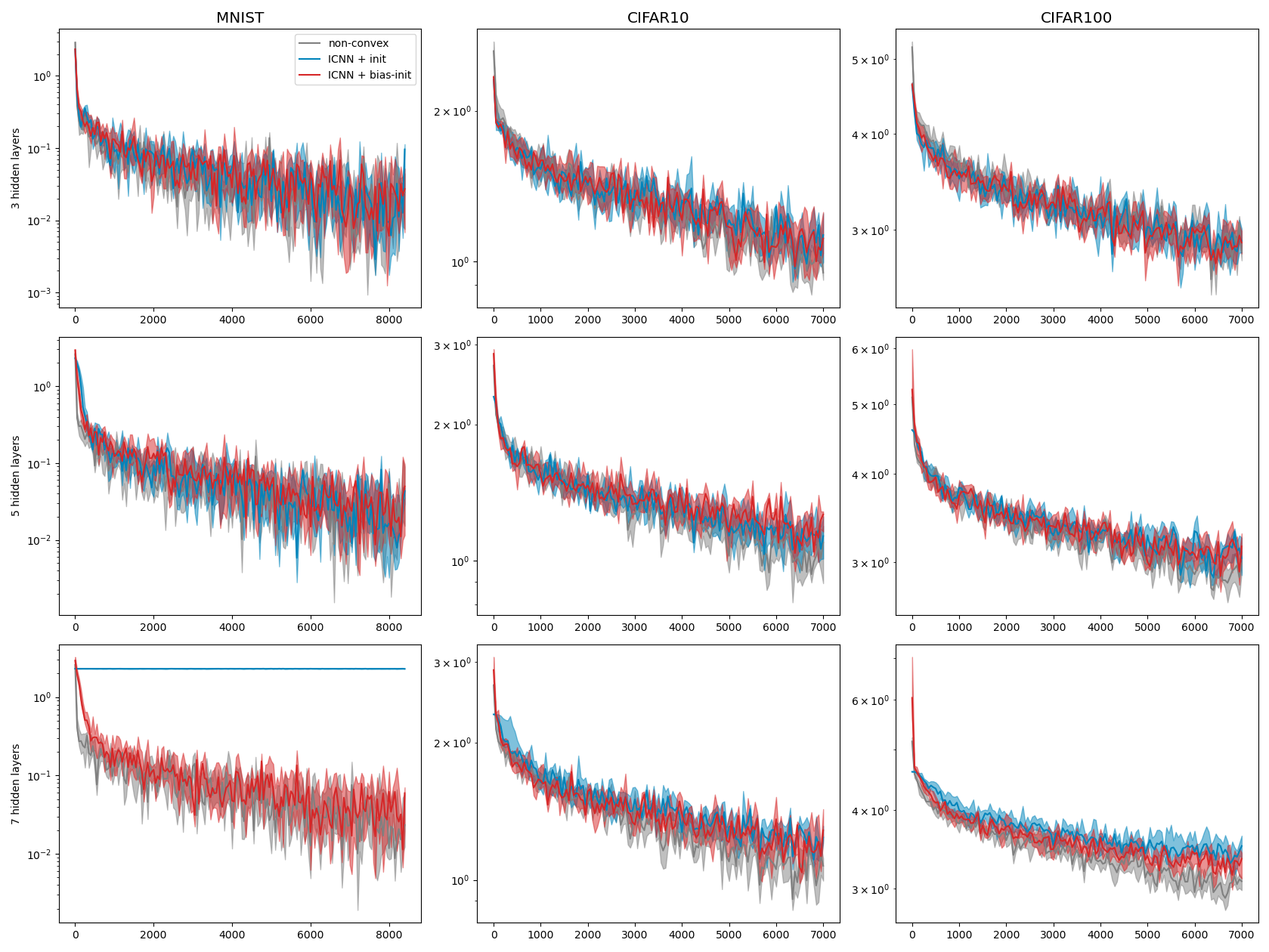}
    \caption{
        Training loss curves of \ac{icnn} variants with the same architecture with three, five and seven hidden layers on the {MNIST}, {CIFAR10} and {CIFAR100} datasets. 
        ``\acs{icnn}~+~init'': our principled initialisation for \acp{icnn} with $\sigma_b^2 = \beta = 0$.
        ``\acs{icnn}~+~bias-init'': our principled initialisation for \acp{icnn} with $\sigma_b^2 = \beta = \frac{1}{2}$. 
        ``non-convex'': a traditional non-convex network. The median performance over ten runs is displayed, shaded regions represent the inter-quartile range.
    }
    \label{fig:learn_curves:rand_bias}
\end{figure}

To get an idea of the effects of setting $\beta > 0$, we ran an additional experiment with $\beta = \frac{1}{2}$.
By introducing variation in the initial bias parameters, the pre-activations at initialisation can become negative as well.
As a result, we suspect that the distribution of the pre-activations can become more Gaussian-like by increasing $\beta$.
Figure~\ref{fig:propagation_bias} provides a comparison of the propagation with $\beta = 0$ and $\beta = \frac{1}{2}$, showing a subtle, yet visible, effect.
Figure~\ref{fig:learn_curves:rand_bias} shows learning curves for $\beta \in \{0, \frac{1}{2}\}$.
Here, we observe that the random biases enables training of \acp{icnn} in the 7-layer network on {MNIST}.
This indicates that initialising the bias parameters with Gaussian random samples can further improve the model performance.

\subsection{Ablation Studies}
\label{app:experiments:ablations}

\begin{figure}
    \centering
    \includegraphics[width=\linewidth]{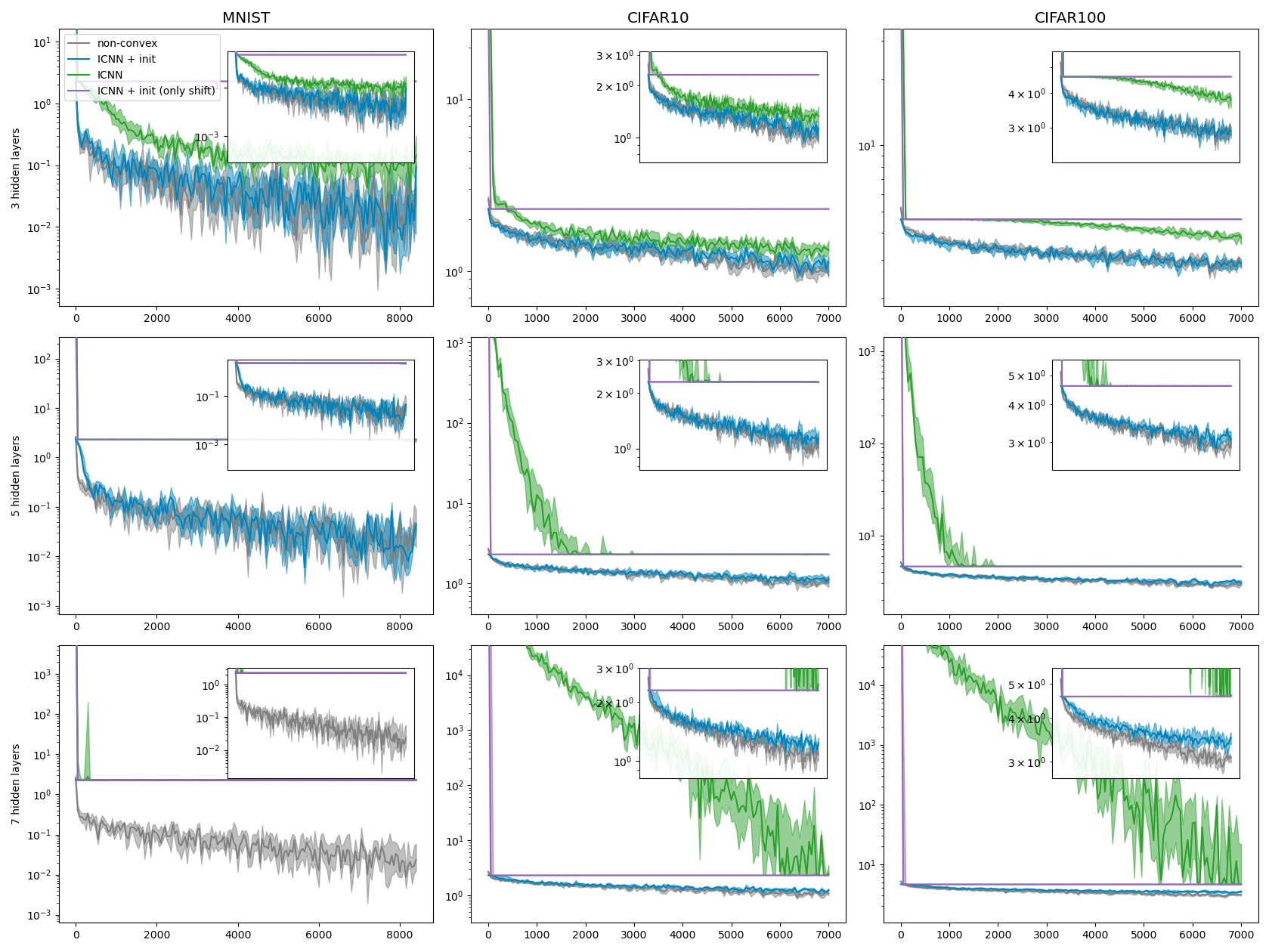}
    \caption{
        Training loss curves of \ac{icnn} variants with the same architecture with three, five and seven hidden layers on the {MNIST}, {CIFAR10} and {CIFAR100} datasets. 
        ``\acs{icnn}~+~init'': our principled initialisation for \acp{icnn} w/o skip-connections.
        ``\acs{icnn}'' is an \acl{icnn} with He initialisation.
        ``\acs{icnn}~+~init (only shift)'': our principled initialisation for \acp{icnn} for bias parameters only. 
        ``non-convex'': a traditional non-convex network. 
        The median performance over ten runs is displayed, shaded regions represent the inter-quartile range.
        Note that learning curves for ``\acs{icnn}'' and ``\acs{icnn}~+~init (only shift)'' overlap on {MNIST}.
    }
    \label{fig:learn_curves:bias_only}
\end{figure}

To better understand the effects of the various hyper-parameters of our proposed method, we ran a set of ablation experiments.
One of the most important factors of our proposed method is the initialisation of the bias parameters.
This raises the question whether it might suffice to initialise the bias parameters (cf.~eq.~\ref{eq:bias:mean}) and ignore the weights entirely.
Figure~\ref{fig:learn_curves:bias_only} compares networks trained with and without our full initialisation to those where only the bias parameters are initialised according to eq.~\eqref{eq:bias:mean}.
The weight parameters for the latter networks are initialised using strategies for regular networks.
Our experiments show that only centring the pre-activations does not suffice to make these networks learn better.
The networks converge faster, but they get stuck on the same plateau that \acp{icnn} that do not use our principled initialisation get stuck on.
For shallow networks, the bias shift even seems to prevent \acp{icnn} from getting past this plateau, leading to inferior learning (first row of Figure~\ref{fig:learn_curves:bias_only}).

\begin{figure}
    \centering
    \includegraphics[width=\linewidth]{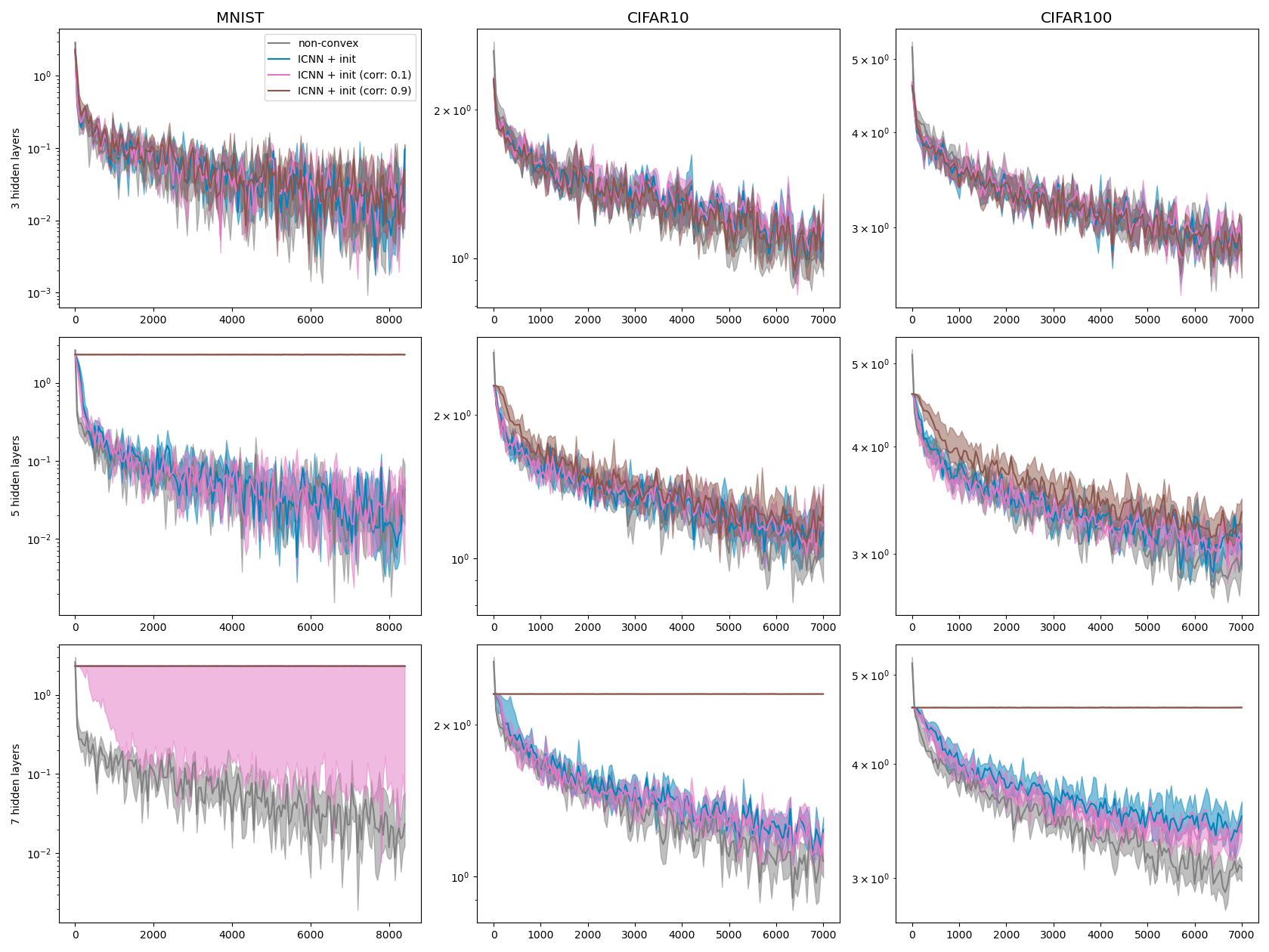}
    \caption{
        Training loss curves of \ac{icnn} variants with the same architecture with three, five and seven hidden layers on the {MNIST}, {CIFAR10} and {CIFAR100} datasets. 
        ``\acs{icnn}~+~init'': our principled initialisation for \acp{icnn} with $\rho_* = \frac{1}{2}$.
        ``\acs{icnn}~+~init (corr: $r$)'': our principled initialisation for \acp{icnn} with $\rho_* = r$. 
        ``non-convex'': a traditional non-convex network. The median performance over ten runs is displayed, shaded regions represent the inter-quartile range.
    }
    \label{fig:learn_curves:corr}
\end{figure}

The choice for $\rho_* = \frac{1}{2}$ in section~\ref{sec:icnn_init} is rather arbitrary.
The main motivation for this choice is that it leads to a closed-form expression for the $\relu$ kernel (see section~\ref{app:sigprop:kernels}).
However, other choices for $\rho_*$ are still possible.
Since the fixed point for $\rho_*$ is not stable (see section~\ref{app:icnn_init:fixpoint_stability}), the correlation still tends to drift towards one.
As a result, choosing a lower value for $\rho_*$ might allow to train even deeper networks.
We verified this experimentally and the learning curves for these experiments can be found in Figure~\ref{fig:learn_curves:corr}.

\subsection{Different ICNN Implementations}
\label{app:experiments:reparam}

\begin{figure}
    \centering
    \includegraphics[width=\linewidth]{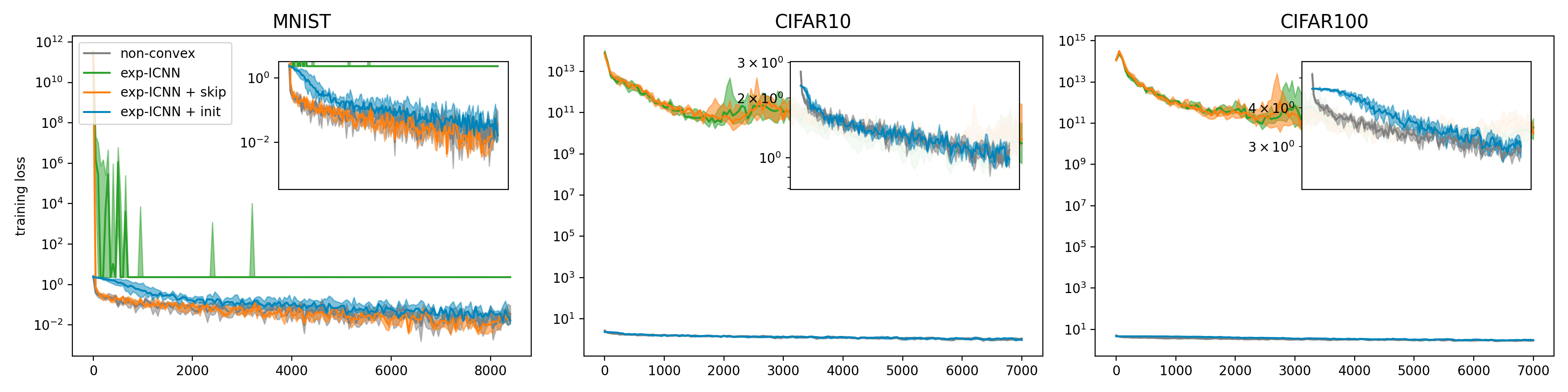}
    \caption{
        Training loss curves of \ac{icnn} variants using an exponential reparameterisation of weights.
        The reference exp-\ac{icnn} has no skip-connections and is initialised with the default He-initialisation.
        The exp-\ac{icnn} + skip model additionally includes skip-connections, but no principled initialisation.
        Our principled initialisation for exp-\acp{icnn} without skip-connections is denoted by \ac{icnn} + init.
        The results for a non-convex network are included for reference.
        Each curve displays the median performance over ten runs.
        Shaded regions represent the region between the quartiles and dashed lines represent min- and maxima over the ten runs.
        The inset figures provide a view of the loss curves zoomed in on the reference non-convex network.
    }
    \label{fig:learn_curves_exp}
\end{figure}

The original \ac{icnn} uses a projection method after every update to keep the weights positive \citep{amos17icnn}.
An alternative method to keep weights positive is to re-parameterise the weights using a function with positive co-domain \citep[cf.][]{nesterov22learning}.
If weights are re-parameterised using the exponential function, we do not need to draw weights from a log-normal distribution.
I.e.~such that $\mat{W} = \exp(\tilde{\mat{W}}),$ where $\exp(\cdot)$ is applied element-wise.
Instead, the initial weights can be directly sampled from a Gaussian distribution. 
Figure~\ref{fig:learn_curves_exp} shows the result for the initialisation experiments from Figure~\ref{fig:learn_curves} using the re-parameterisation instead of the projection method.

\begin{figure}
    \centering
    \begin{subfigure}{.49\linewidth}
        \includegraphics[width=\linewidth]{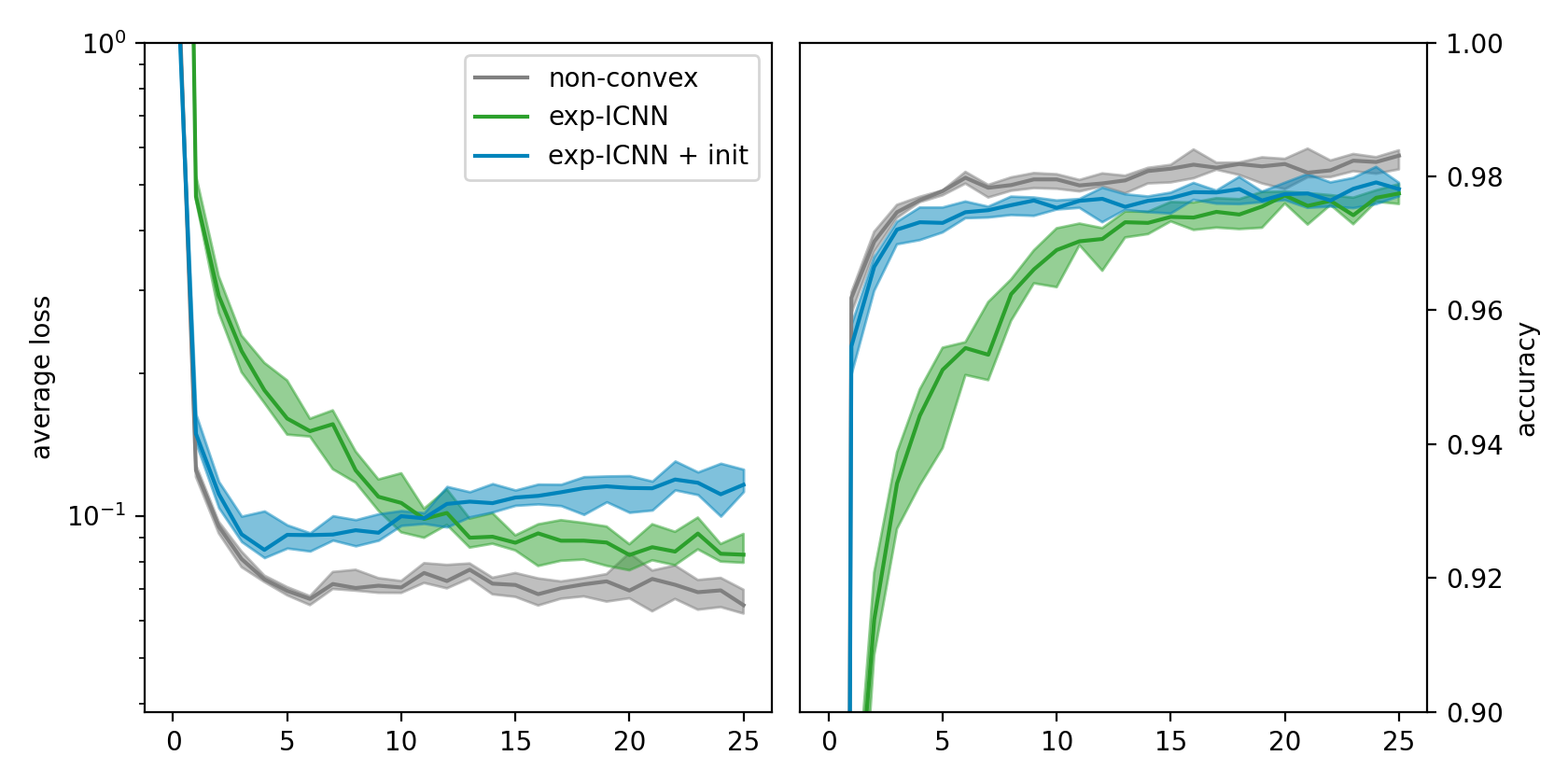}
        \caption{MNIST}
    \end{subfigure}%
    \begin{subfigure}{.49\linewidth}
        \includegraphics[width=\linewidth]{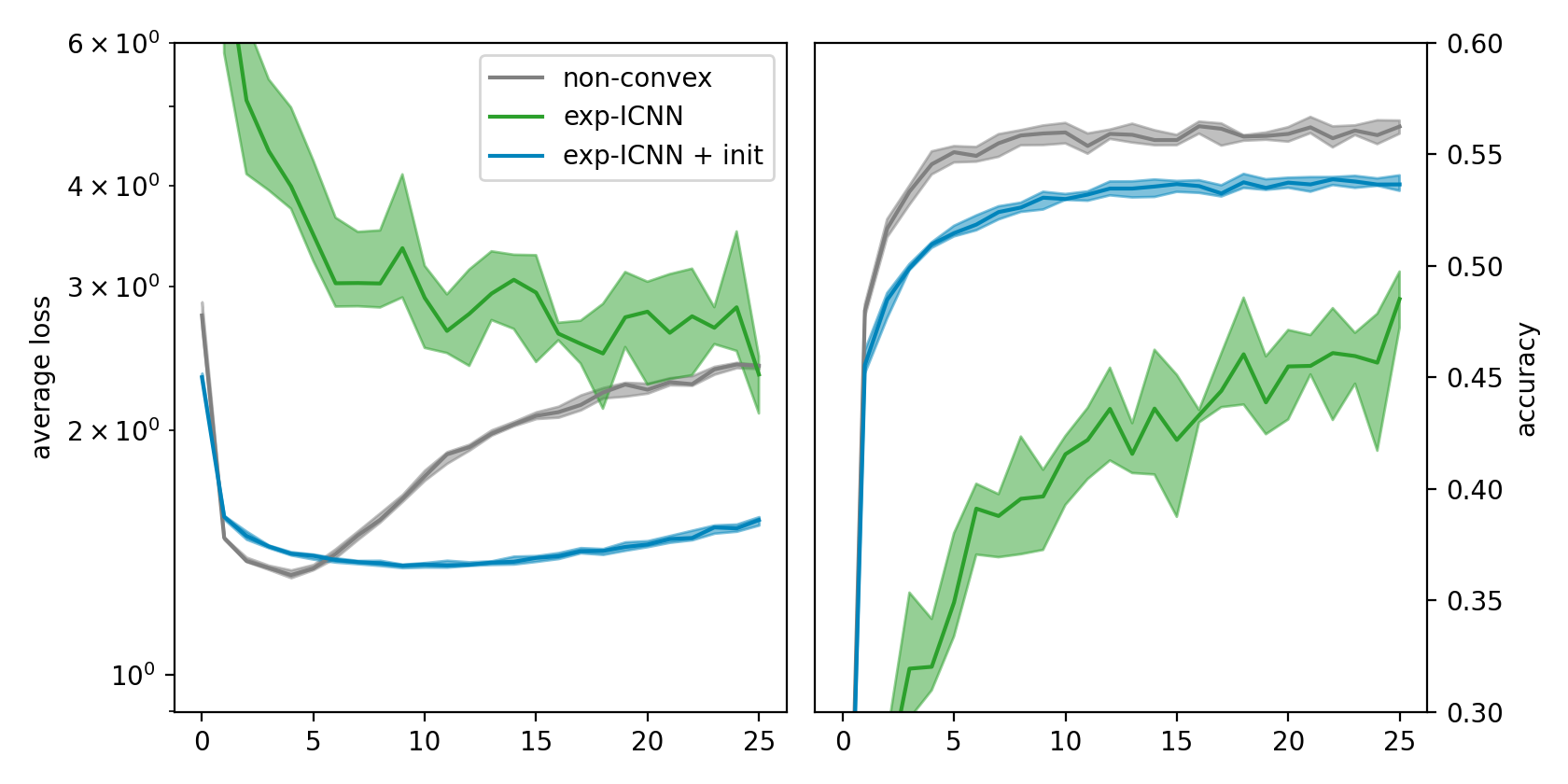}
        \caption{CIFAR10}
    \end{subfigure}
    \caption{
        Test set metrics of compared methods on the (a) {MNIST} and (b) {CIFAR10} datasets. 
        Each curve displays the median performance over ten runs.
        Shaded regions represent the inter-quartile range over the ten runs.
    }
    \label{fig:fine_tuned_exp}
\end{figure}

Figure~\ref{fig:learn_curves_exp} shows that our initialisation makes it possible to train \acp{icnn} using the exponential re-parameterisation.
This re-parameterisation greatly affects the learning dynamics of the network, however.
Therefore, we find that this particular implementation of \acp{icnn} generally does not perform quite as well as the original implementation.
Figure~\ref{fig:fine_tuned_exp} shows the learning curves for re-parameterised \acp{icnn} using the hyper-parameters from table~\ref{tab:exp_hparams}.
The search space (in Table~\ref{tab:hparam_space}) is the same as for Figure~\ref{fig:fine_tuned} in the main paper.

\begin{table}
    \centering
    \caption{
        Optimal hyper-parameters for each configuration in Figure~\ref{fig:fine_tuned_exp}. 
        The epoch column reports the epoch in which the validation accuracy was highest.
    }
    \label{tab:exp_hparams}
    \begin{tabular}{lllccccr}
        \toprule
        &           & architecture                & $\eta$ & $L_2$ & pre-processing & epoch & accuracy \\ \midrule
        \multirow{3}{1em}{\rotatebox{90}{MNIST}} 
        & non-convex  & (784, 1\,000, 1\,000, 10)          & 1e-4   & 0.01  & {ZCA}   & 25 & 98.62\%  \\
        & exp-\ac{icnn} & (784, 1\,000, 10)                & 1e-4   & 0.01  & none    & 25 & 98.02\%  \\
        & exp-\ac{icnn} + init & (784, 1\,000, 1\,000, 10) & 1e-2   & 0.00  & {ZCA}   & 25 & 98.18\%  \\
        \midrule
        \multirow{3}{1em}{\rotatebox{90}{CIFAR}}
        & non-convex  & (3072, 10\,000, 1\,000, 10)  & 1e-4   & 0.01  & {ZCA}    & 18 & 55.92\%  \\
        & exp-\ac{icnn} & (3072, 1\,000, 10)         & 1e-4   & 0.01  & none     & 22 & 48.38\% \\
        & exp-\ac{icnn} + init & (3072, 10\,000, 10) & 1e-3   & 0.00  & {PCA}    & 24 & 55.10\%  \\
        \bottomrule
    \end{tabular}
\end{table}

\end{document}